\documentclass[journal]{IEEEtran}
\usepackage{cite}
\usepackage{amsmath,amssymb,amsfonts}
\usepackage{url}
\usepackage{amsmath}
\usepackage{algorithmic}
\usepackage{graphicx}
\usepackage{textcomp}
\usepackage{algorithm}
\usepackage{booktabs}
\usepackage{tabularx}
\usepackage{multirow}
\usepackage{arydshln}
\usepackage{array}
\usepackage{ragged2e}
\usepackage{adjustbox}
\usepackage{hyperref}

\begin{document}
\title{Dual-Reference Source-Free Active Domain Adaptation for Nasopharyngeal Carcinoma Tumor Segmentation across Multiple Hospitals
}
\author{Hongqiu Wang, Jian Chen, Shichen Zhang, Yuan He, Jinfeng Xu, Mengwan Wu, Jinlan He, \\Wenjun Liao, Xiangde Luo
\vspace{-10mm}
\thanks{This work was supported by the National Natural Science Foundation of China under Grant 82203197. Corresponding author: Xiangde Luo (xiangde.luo@std.uestc.edu.cn).}
\thanks{H. Wang and S. Zhang are with the Department of Systems Hub, Hong Kong University of Science and Technology (Guangzhou), Guangzhou 511400, China.}
\thanks{J. Chen is with the Department of Radiology, University of Cambridge, Cambridge CB2 1TN, UK.}
\thanks{Y. He is with the Department of Radiation Oncology, Anhui Provincial Hospital, University of Science and Technology of China, Hefei 23000, China.}
\thanks{J. Xu is with the Department of Radiation Oncology, Nanfang Hospital, Southern Medical University, Guangzhou 510515, China.} 
\thanks{M. Wu is with the Cancer Center, Sichuan Provincial People’s Hospital, Chengdu 610041, China.}
\thanks{J. He is with the Department of Radiation Oncology, West China Hospital, Sichuan University, Chengdu 610041, China.}
\thanks{W. Liao and X. Luo are with the Department of Radiation Oncology, Sichuan Cancer Hospital and Institute, University of Electronic Science and Technology of China, Chengdu 610072, China. X. Luo is also with Shanghai Artificial Intelligence Laboratory,
Shanghai 200030, China.}}

\maketitle
\begin{abstract}
Nasopharyngeal carcinoma (NPC) is a prevalent and clinically significant malignancy that predominantly impacts the head and neck area. Precise delineation of the Gross Tumor Volume (GTV) plays a pivotal role in ensuring effective radiotherapy for NPC. Despite recent methods that have achieved promising results on GTV segmentation, they are still limited by lacking carefully-annotated data and hard-to-access data from multiple hospitals in clinical practice. Although some unsupervised domain adaptation (UDA) has been proposed to alleviate this problem, unconditionally mapping the distribution distorts the underlying structural information, leading to inferior performance. To address this challenge, we devise a novel Sourece-Free Active Domain Adaptation (SFADA) framework to facilitate domain adaptation for the GTV segmentation task. Specifically, we design a dual reference strategy to select domain-invariant and domain-specific representative samples from a specific target domain for annotation and model fine-tuning without relying on source-domain data. Our approach not only ensures data privacy but also reduces the workload for oncologists as it just requires annotating a few representative samples from the target domain and does not need to access the source data. We collect a large-scale clinical dataset comprising 1057 NPC patients from five hospitals to validate our approach. Experimental results show that our method outperforms the UDA methods and achieves comparable results to the fully supervised upper bound, even with few annotations, highlighting the significant medical utility of our approach. In addition, there is no public dataset about multi-center NPC segmentation, we will release code and dataset for future research: \url{https://github.com/whq-xxh/Active-GTV-Seg}.
\end{abstract}

\section{Introduction}
\label{sec:introduction}
\IEEEPARstart{N}{asopharyngeal} carcinoma (NPC) is a prevalent malignancy affecting the head and neck region, and Intensity-Modulated Radiation Therapy (IMRT) has emerged as a preferred radiation technique for its treatment \cite{chua2016nasopharyngeal,lee2002intensity,luo2023deep}. IMRT has demonstrated noticeable advancements in enhancing the 5-year locoregional control rate and reducing radiation-associated toxicities among NPC patients \cite{chen2019nasopharyngeal}. Accurate delineation of the GTV based on Magnetic Resonance Imaging (MRI) is critical in radiation therapy, particularly in IMRT for NPC \cite{kam2004treatment,razek2012mri}. However, GTV contouring for NPC poses significant challenges due to its complex anatomical structures and diverse tumor invasion pathways. Manual GTV delineation is time-consuming, labor-intensive, and prone to errors and inter-observer variability \cite{lin2019deep}. These factors can adversely impact the accuracy of GTV contouring and potentially result in treatment failures \cite{chen2022failure}. 

Recently, the application of deep learning techniques has substantially advanced the field of medical auto-delineation, yielding successful models for various cancer types\cite{wang2020boundary,jin2021deeptarget,tian2023delineation}, including the NPC \cite{li2022npcnet}. These models have demonstrated highly promising segmentation outcomes, showcasing their potential for precise and efficient delineation in clinical practice.

Despite the remarkable progress in GTV segmentation methods and the availability of data for NPC, there are still several challenges that hinder the widespread clinical adoption of deep learning techniques in this domain. Firstly, GTV segmentation tasks typically require a large amount of well-annotated MRI training samples, which necessitates extensive manual effort and time-consuming annotation processes. Secondly, deep models often lack generalizability across different data sources, as variations in equipment vendors, imaging resolution, slice thickness and delineation styles among medical centers can impact model performance. Thirdly, strict data privacy regulations and diverse information systems in clinical settings limit data sharing, posing a challenge for transfer learning and domain adaptation approaches that rely on access to both source and target domain annotated data \cite{weiss2016survey,wang2020tent}. Addressing these challenges is paramount in facilitating the practical application of deep learning-based GTV segmentation methods for NPC radiotherapy.

\begin{figure}[h]
    \centering
    \includegraphics[width=0.5\textwidth]{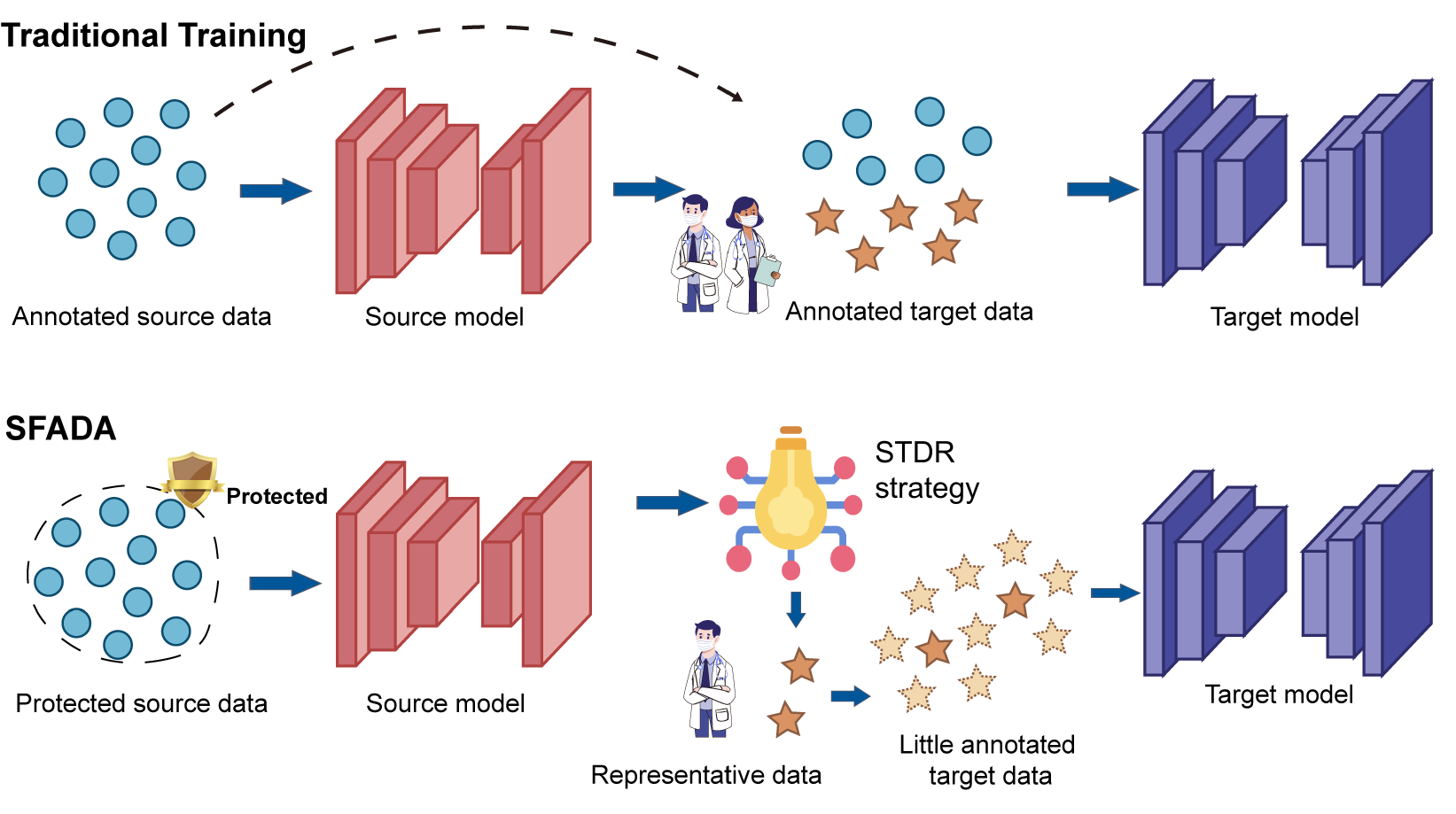}
    \vspace{-6mm}
    \caption{Comparison of traditional training and our Source-Free Active Domain Adaptation (SFADA) training. Our approach safeguards the source data while demanding only a minimal annotation effort. }
    \label{fig:F1}
    \vspace{-3mm}
\end{figure}

We acknowledge the existence of UDA methods that partially address the aforementioned challenges. However, based on our primary experiments and research, we have found that while applying widely investigated UDA methods does yield some improvement compared to directly using a source domain model without fine-tuning, this improvement is quite limited. Notably, there still exists a substantial performance gap between UDA methods and fully supervised trained models. This considerable gap greatly reduces the practical value of computer-aided segmentation. Hence, there is a pressing need to enhance the effectiveness of domain adaptation techniques in medical image segmentation by exploring more practical and feasible approaches that can effectively bridge this performance gap.

In this paper, we pioneer the introduction of a novel approach called Source-Free Active Domain Adaptation for cross-domain GTV segmentation across different medical centers. This method is specifically designed to address the unique requirements and challenges in the field of medicine as shown in Fig.~\ref{fig:F1}. To begin with, our approach eliminates the need for accessing the source data, ensuring strict data privacy compliance and adherence to hospital regulations. Meanwhile, it reduces the complexity of data transmission, resulting in significant time and resource savings. Additionally, our method facilitates knowledge transfer of model segmentation across diverse healthcare domains, promoting seamless collaboration and improving overall efficiency. Another pivotal aspect of our approach is the utilization of the proposed Source-domain and Target-domain Dual-Reference (STDR) strategy, enabling the active selection of representative domain-invariant and domain-specific samples for recommendation to radiologists for annotation. It is important to highlight that this strategy is highly practical, as these selected samples constitute only a small fraction of the overall dataset. This approach effectively reduces the workload and saves valuable time for radiologists, while enabling radiologists to concentrate their efforts on the most representative and information-rich key samples for segmentation. Furthermore, subsequent experimental results demonstrate that our method attains comparable outcomes to those achieved through fully supervised training in the target domain. This underscores the remarkable practical value of our approach and its potential for real-world applications.

To substantiate the efficacy of our approach, we assemble a comprehensive dataset comprising MRI scans from 1057 NPC patients, sourced from five diverse medical centers for experiments on cross-domain GTV segmentation. Additionally, we conducted a thorough comparison with state-of-the-art (SOTA) domain adaptation methods (both with and without access to source data), as well as active learning methods to demonstrate the superiority of our method. The main contributions could be summarized as follows: 

\begin{itemize}
\item To the best of our knowledge, We are the first to propose the Source-Free Active Domain Adaptation (SFADA) method in the medical field. We also verify the effectiveness of SFADA on cross-domain GTV segmentation. SFADA provides a new feasible solution for the practical implementation of computer-aided GTV segmentation, and remarkably improves the segmentation accuracy.
\item We propose an STDR to actively select representative samples for annotation and further reduce label costs without relying on source-domain data.
\item We employ a semi-supervised learning strategy, leveraging both unlabeled and labeled samples chosen through our STDR strategy, to jointly train the model. This allows the model to learn from more diverse examples, therefore, enhancing the segmentation accuracy.
\item Our approach outperforms other SOTA domain adaptation methods on five hospitals' clinical NPC datasets. Notably, the proposed method with a few labeled representative samples can match the performance of fully supervised training on the target domain. Moreover, we build the first multi-center NPC GTV dataset for public research.
\end{itemize}

\section{Related work}
\subsection{GTV Segmentation in NPC}
In recent years, there have been numerous studies focusing on the segmentation of NPC \cite{li2022npcnet}. These studies can be categorized into two main approaches: conventional machine learning-based methods and deep learning-based methods. Zhou \textit{et al.} \cite{zhou2006nasopharyngeal} introduce a segmentation technique for NPC based on a two-class support vector machine (SVM). This method aims to find an optimal hyperplane to effectively separate the different classes in the MR images. Huang \textit{et al.} \cite{huang2013region} present two region-based methods, the metric-based similarity learning and the discriminative classification-based with kernel learning for NPC segmentation. Men \textit{et al.} \cite{men2017deep} propose an end-to-end deep deconvolutional neural network to achieve pixel segmentation in CT images for NPC patients. Chen \textit{et al.} \cite{chen2020mmfnet} propose a multi-encoder single-decoder network for accurate segmentation of NPC using multi-modal MRI data. Liao \textit{et al.} \cite{liao2022automatic} evaluate the clinical value of a semi-supervised learning framework for GTV delineation in NPC, demonstrating its accuracy and potential to assist oncologists in improving contouring accuracy and reducing contouring time. Li \textit{et al.} \cite{li2022npcnet} propose a comprehensive framework for primary NPC tumor segmentation, incorporating position enhancement, scale enhancement, and boundary enhancement modules. The results demonstrate the effectiveness of the approach in achieving accurate segmentation outcomes. While great progress has been made in NPC segmentation, effective methods for cross-center GTV segmentation remain lacking. Hence, our study focuses on leveraging segmentation knowledge from the source domain to the target domain, enabling comparable performance with minimal annotated target domain data.

\subsection{Domain adaptation}

\vspace{3pt}\noindent\textbf{Unsupervised Domain Adaptation:} UDA is commonly trained using labeled data from the source domain and unlabeled data from the target domain. It relies on the distributional discrepancies between the source and target domain data to achieve domain adaptation through adversarial training and minimizing inter-domain distance \cite{wilson2020survey}. 
Tsai \textit{et al.} \cite{tsai2018learning} leverage the structured property of pixel-level predictions, which encode spatial and local information, to propose an effective domain adaptation algorithm based on adversarial learning in the output space. 
Vu \textit{et al.} \cite{vu2019advent} introduce a novel approach that maximizes prediction certainty in the target domain, improving UDA performance. It utilizes an entropy-based adversarial training method to minimize entropy and facilitate structural adaptation from the source domain to the target domain.

\vspace{3pt}\noindent\textbf{Source-Free Domain Adaptation:} Source-free domain adaptation (SFDA) is a technique that enables model adaptation to target domains without relying on any source domain data. It transfers knowledge and adapts models to target domains, addressing data availability and privacy concerns. This approach is particularly valuable in scenarios where source data is scarce or inaccessible, such as in medical settings \cite{guan2021domain}. 
Wang \textit{et al.} \cite{wang2020tent} introduce a confidence optimization technique that leverages entropy-based model predictions for domain adaptation without requiring access to the source data. By estimating normalization statistics and performing online transformations, the method enhances model confidence and reduces uncertainty. 
Fleuret \textit{et al.} \cite{fleuret2021uncertainty} propose an approach to mitigate prediction uncertainty in target domain data by focusing on two aspects: minimizing the entropy of the predicted posterior and maximizing the robustness of the feature representation against noise. 
Chen \textit{et al.} \cite{chen2021source} propose a denoised pseudo-labeling approach for SFDA. This method improves the quality of supervision by incorporating pixel-level and class-level denoising schemes. It enables efficient adaptation to the target domain without relying on source data.

\subsection{Active learning}
Active learning is a strategy that aims to achieve optimal performance while minimizing annotation costs. It makes efficient utilization of the annotation budget by focusing on the most valuable samples for improving performance \cite{cohn1996active}. Various strategies have been developed for sample selection. These strategies include uncertainty-based methods that focus on selecting samples with high uncertainty \cite{lewis1994heterogeneous,scheffer2001active}, diversity-based methods that aim to maximize the diversity of selected samples \cite{jain2016active,hoi2009semisupervised}, representativeness-based methods that prioritize samples representing different regions of the data distribution \cite{xu2003representative,huang2010active}, and strategies that aim to maximize the expected label changes \cite{freytag2014selecting,kading2015active}. These diverse approaches provide different perspectives on selecting informative samples in active learning and these methods have been widely utilized in computer vision tasks, such as image classification and image segmentation. In this study, we propose the integration of active learning into domain adaptation to mitigate the distortion of the target-domain distribution. Active learning offers the advantage of minimal annotation cost, making it feasible in many medical scenarios considering the potential performance improvements.

Currently, the application of active learning to the domain adaptation problem remains relatively limited. Su \textit{et al.} \cite{su2020active} introduce an Active Adversarial Domain Adaptation (AADA) strategy, which combines domain adversarial learning and active learning. AADA takes into account both the uncertainty and diversity of the samples during the selection process, reducing labeling costs on the target domain on object classification and detection tasks. However, this adversarial-based learning approach relies on access to source data. In this paper, we introduce a novel Source-Free Active Domain Adaptation method specifically designed for medical scenarios, addressing the challenge of cross-center medical image segmentation.

\section{Methodology}
\begin{figure*}[h]
    \centering
    \includegraphics[width=1\textwidth]{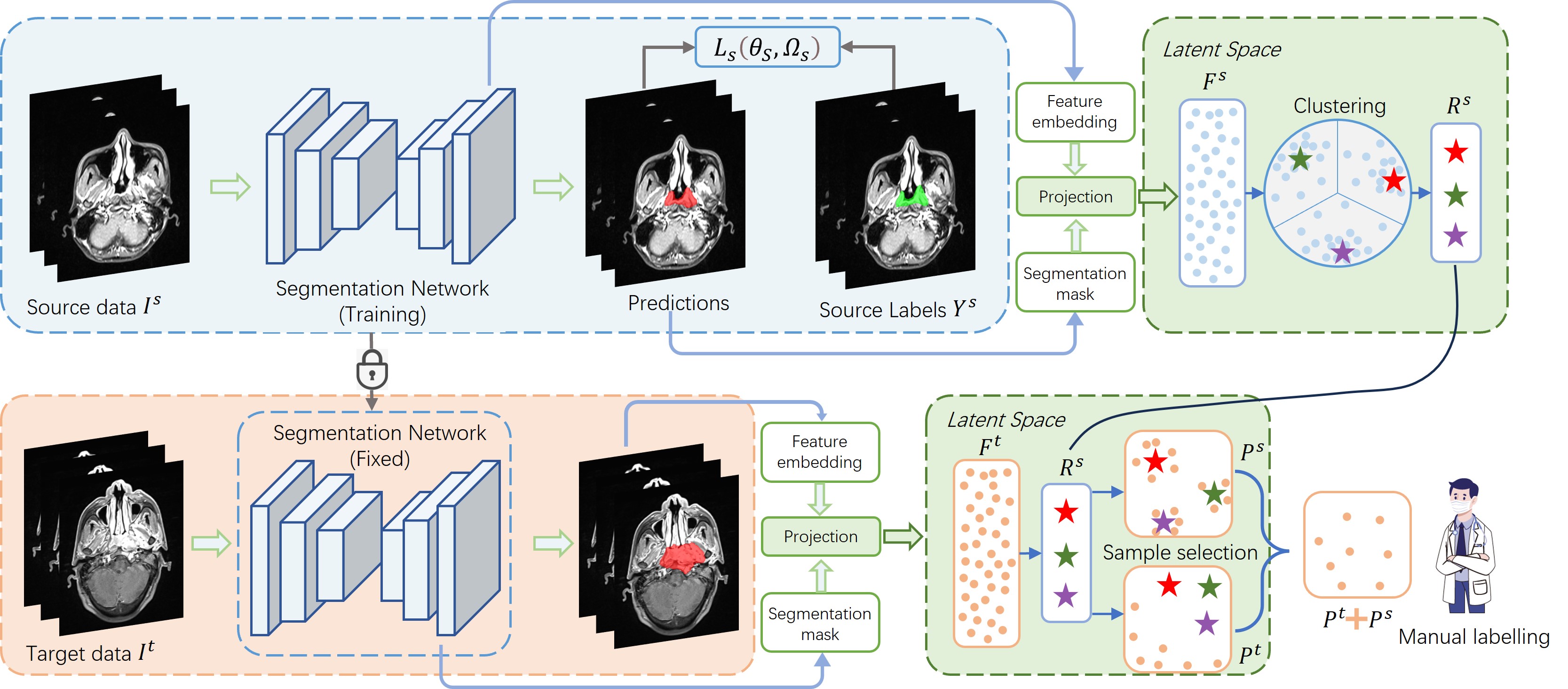}
    \caption{Overview of the STDR strategy. Given the source MR images $I^s$ accompanied by its ground truth masks $Y^s$, our STDR trains the segmentation network in the source domain firstly. The segmented features are then mapped to the latent space as $F^s$, and clustering algorithms derive the reference point $R^s$. The data in the target domain undergoes similar manipulation to obtain $F^t$. We then select two types of representative samples for manual labelling by radiologists.}
    \label{fig:STDR}
    \vspace{-4mm}
\end{figure*}

\begin{figure}[h]
    \centering
    \includegraphics[width=0.5\textwidth]{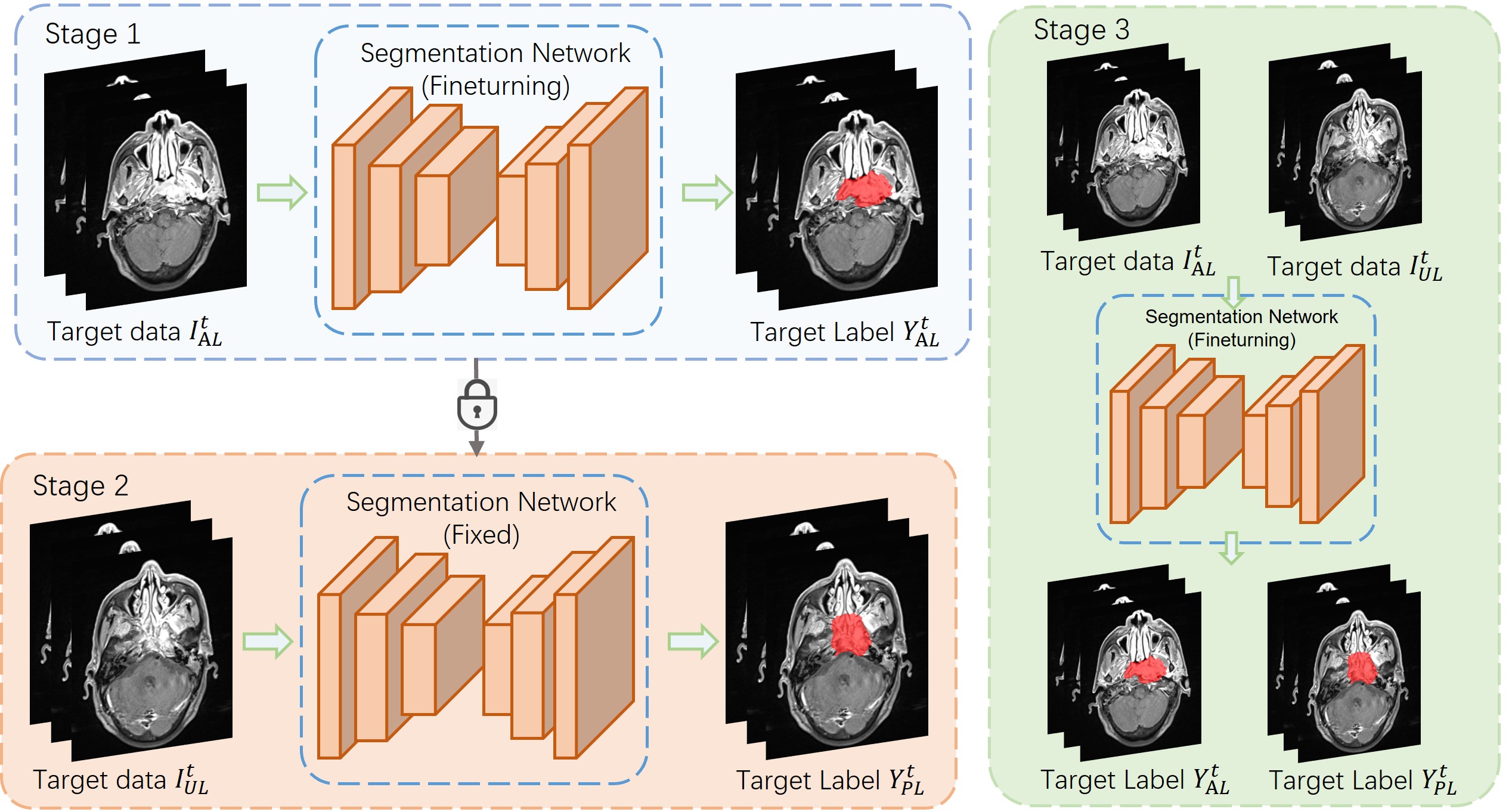}
    \vspace{-5mm}
    \caption{Detailed pipeline of the proposed semi-supervised method. The whole process consists of three stages.}
    \label{fig:semi}
\end{figure}

In this chapter, we begin by presenting the construction of our cross-center GTV segmentation dataset. Subsequently, we provide a detailed description of our Source-domain and Target-domain Dual-Reference (STDR) strategy (as shown in Fig.~\ref{fig:STDR}), which assists in selecting the most representative samples and helps us to learn about domain-invariant and domain-specific knowledge. In addition, Fig.~\ref{fig:semi} describes the semi-supervised workflow. This allows the network to effectively utilize the abundant unlabeled data and improve the generalization capability of the model. Finally, we provide details on the training and implementation of our model.

\subsection{Dataset Construction}

\begin{figure}[h]
    \centering
    \includegraphics[width=0.46\textwidth]{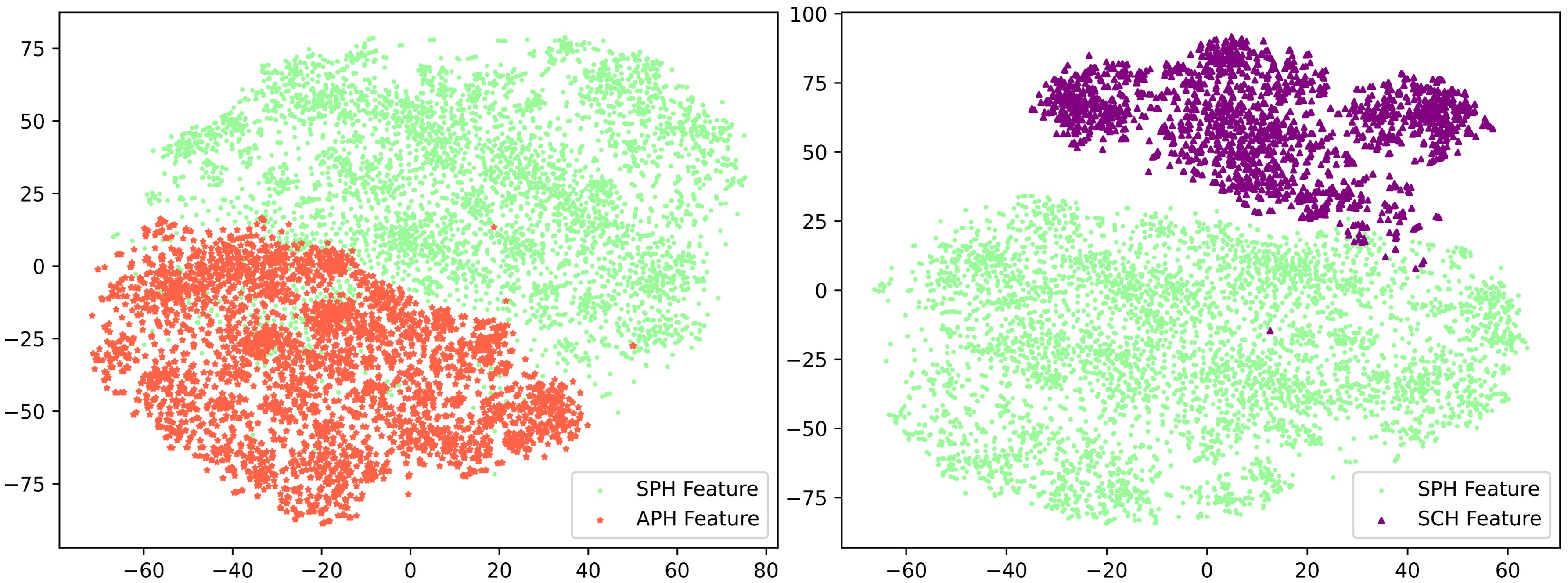}
    \vspace{-1mm}
    \caption{Visualization (t-SNE \cite{hinton2002stochastic}) of the multiple medical centres distribution. The green dot and the orange star and purple triangle represent the latent representation of SPH and APH and SCH respectively.}
    \label{fig:tsne}
    \vspace{-3mm}
\end{figure}

\begin{table}[h]
\centering
\caption{The dataset comprises data acquired from five diverse medical centers, including information on the volume, vendors and Magnetic field of the data.}
\vspace{-1mm}
\label{tab1}
\resizebox{0.48\textwidth}{!}{%
\begin{tabular}{c|c|c|c}
\hline
Data source& Patients  & Vendors (Patients) & Magnetic field (Patients)   \\
\hline
Sichuan Cancer Hospital (SCH) & 52 & Siemens (52)& 1.5T (23), 3T (29)   \\
Anhui Provincial Hospital (APH) & 146 & GE (113), Siemens (27), Philips (6)& 1.5T (119), 3T (27) \\
Sichuan Provincial People's Hospital (SPH) & 208 & Siemens (208)& 1.5T (10), 3T (198)  \\
West China Hospital (WCH) & 284 & GE (284)&3T (284) \\
Sothern Medical University (SMU) & 367& GE (334), Siemens (24), Philips (9)& 1.5T (336), 3T (31) \\
\hline
\end{tabular}}
\vspace{-3mm}
\end{table}

We acquire the NPC MRI datasets with annotated labels from five renowned medical institutions, focusing specifically on patients diagnosed with NPC. The MR images are obtained by utilizing different scanners with 1.5-T or 3.0-T, with a large range of inter-slice thickness (range, 1.0-8.0 mm) and a middle range in-plane spacing (range, 0.47-1.67 mm). Strict selection criteria are applied, including histological confirmation of NPC and prior MRI evaluations of the nasopharynx and neck before initiating anticancer treatments. The MRI sequences of this dataset consisted of contrast-enhanced T1-weighted sequences as well as unenhanced T1- and T2-weighted sequences. But to reduce the impact of multi-sequence imaging, we just use the T1-weighted images to investigate the performance of different methods in this study~\cite{luo2023deep}. The basic information regarding the data distribution and vendors of the multi-center NPC MRI datasets is provided in Table~\ref{tab1}. Notably, the t-SNE visualization \cite{hinton2002stochastic} depicted in Fig.~\ref{fig:tsne} demonstrates significant variations in the distributions of the different medical center domains. Our primary objective is to overcome potential domain shift challenges and achieve accurate GTV delineations for patients within the designated medical center.

\subsection{Source-domain and Target-domain Dual-Reference}
\begin{figure}[t]
    \centering
    \includegraphics[width=0.46\textwidth]{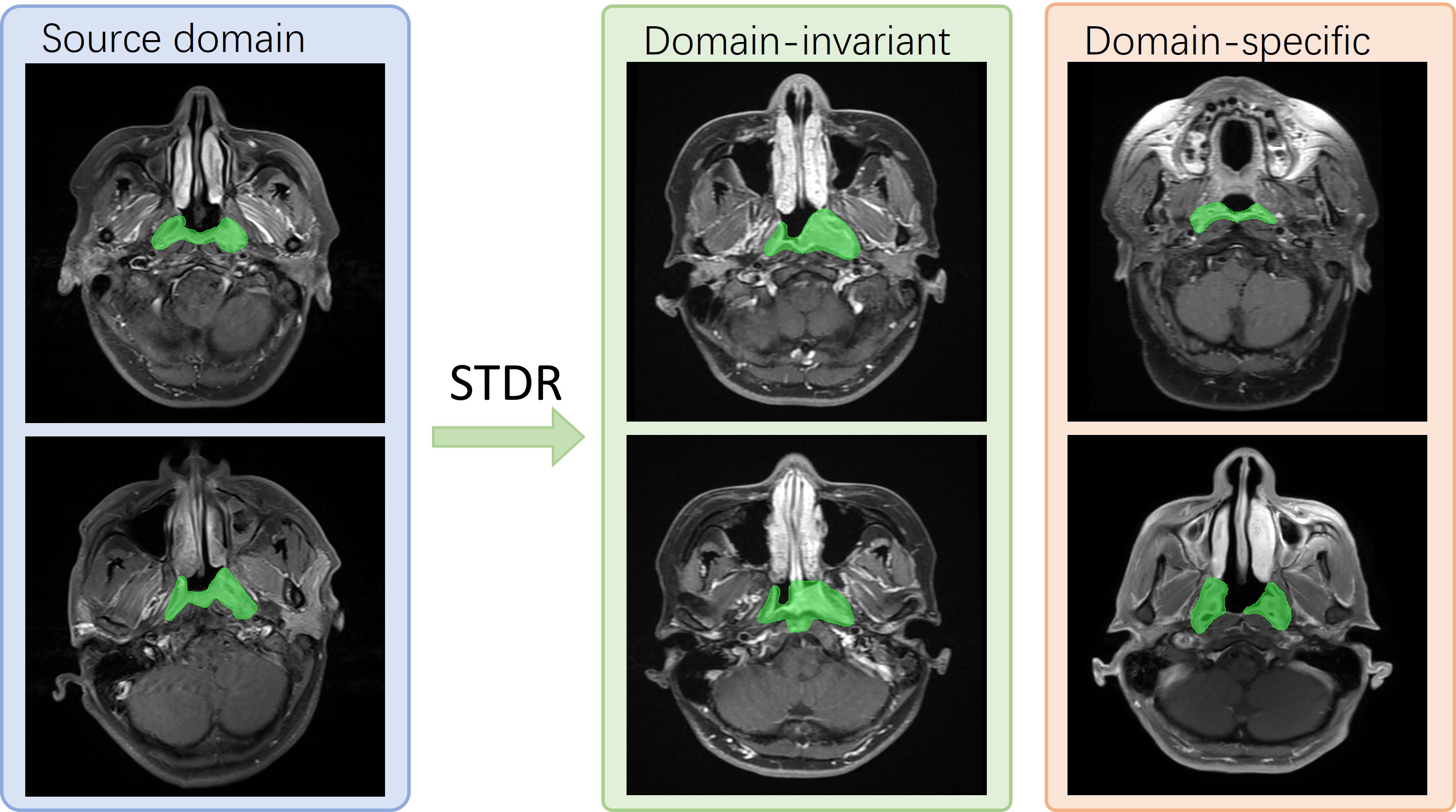}
    \vspace{-1mm}
    \caption{Examples of source domain (SPH), domain-invariant and domain-specific samples (APH). The source domain images present higher similarity with the domain-invariant images,  while the discrepancies with the domain-specific images become more obvious.}
    \label{fig:MRI}
    \vspace{-3mm}
\end{figure}

Fig.~\ref{fig:STDR} presents a schematic illustration of our STDR framework. Our Source-Free Active Domain Adaptation task differs significantly from past active learning for domain adaptation as we lack access to the source data during model fine-tuning. This distinguishes our approach from methods that can be jointly trained using both source and target domain data. In cross-center GTV segmentation, a significant challenge arises from the domain gap in data distribution between different centers, while still sharing some common structural segmentation knowledge. However, learning on a small number of active samples selected from the target data without accessing the source data easily leads to overfitting on the limited samples and potentially causes forgetting of valuable segmentation knowledge gained previously. Therefore, our proposed STDR strategy serves as a reference by jointly considering the data from the source and target domains to select the source domain representative samples $P^s$ and the target domain representative samples $P^t$. This approach offers two key advantages. During fine-tuning in the target domain, samples $P^s$ reinforce the sharing segmentation knowledge without accessing the source data and aid in learning domain-invariant representations. Simultaneously, samples $P^t$ facilitate effective migration of the model to the target domain data, thereby overcoming potential domain shift challenges. Fig.~\ref{fig:MRI} illustrates some examples of MRI images selected by STDR.

The input for the entire framework consists of the following components: the MR images of NPC patients from the source center, denoted as $I^s$: $\Omega_s \subset \mathcal{R}^2 \rightarrow \mathcal{R}$, their respective manual labels $Y^s$ for each pixel $i \in \Omega_s$, where $y_s(i) \in \{0, 1\}$, and the MR images $I^t$ from the target center. 
In the SFDA scenario, we begin by training the segmentation network on source images $I^s$ and their corresponding ground truth labels $Y^s$ in a fully supervised manner. This entails minimizing the loss function with respect to the network parameters $\theta_s$, which is formulated as below: 
\begin{equation}
L_s(\theta_s, \Omega_s) = \frac{1}{|\Omega_s|} \sum_{i=1}^{S} \Psi(y_s(i), p_s(i, \theta_s)). 
\end{equation}
The softmax output of the segmentation network at pixel $i$ in the source image $I^s$ is denoted as $p_s(i, \theta) \in [0, 1]$. We utilize a composite loss function $\Psi$ defined as follows:
\begin{equation}
\small
\begin{aligned}
\Psi(y_s(i), p_s(i, \theta)) &= - \sum_{i=1}^{N} y_{s}(i) \log p_{s}(i, \theta) \\
&\quad + 1 - \frac{2 \sum_{i=1}^{N} y_{s}(i) \hat y_{s}(i)}{\sum_{i=1}^{N} y_{s}(i)^2 + \sum_{i=1}^{N} \hat y_{s}(i)^2},   \label{e2}
\end{aligned}
\end{equation}
Where $\hat y_{s}(i)$ denotes the $i_{th}$ element of the predicted masks, and $N$ represents the overall count of elements.

\vspace{3pt}\noindent\textbf{Dual-Reference Selection.} Considering the presence of multiple radiologists performing GTV outlining and the probable utilization of diverse imaging devices from various vendors in a medical center, it is essential to select multiple reference points $R^s$ to effectively capture this data domain's distributional characteristics. This approach ensures a comprehensive representation of the data and accounts for potential variations introduced by different radiologists and imaging equipment. To achieve this, we initially freeze the above pre-trained segmentation network. Then, we merge the feature embeddings from the penultimate layer of the network with the corresponding predicted masks to jointly map to a latent space. The projection is as follows:
\begin{equation}
 F^s(I^s(i)) =  \frac{1}{\left\| N^s \right\|} MaxP_f(AvgP_c(\hat Y^{s}(i) \otimes f_E(I^s(i)))),   \label{e3}
\end{equation}
where ${\left\| N^s \right\|}$ represents the total number of pixels belonging to the GTV, $\hat Y_{s}(i)$ denotes the $i_{th}$ predicted mask, and $f_E(I^s(i)) \in R^{C\times H\times W}$ represents the output feature embedding from the penultimate layer of $i_{th}$ MRI image. $MaxP_f$ is the max-pooling operation on image features with spatial dimensions $R^{H\times W}$, while $AvgP_c$ is the average pooling operation along the channel dimension $R^{C}$, reducing the feature tensor from three dimensions to two dimensions. Notably, in our dense prediction task, we select a small kernel size in Max pooling to extract features from multiple image patches. This approach provides a more comprehensive representation compared to pooling the entire image into a single element. The final $F^s$ is a flattening of this two-dimensional vector.

Subsequently, we employ the clustering method (i.e., K-means \cite{macqueen1967some}) to group feature vectors from all source images into K clusters, aiming to minimize the following error:
\begin{equation}
\sum_{k=1}^{K} \sum_{i \in R_k} {\left\|F^s(I^s(i)) - R^s_k \right\|}^2_2,
\end{equation}
where ${\left\| \cdot \right\|}^2_2$ refers to the Euclidean distance, and $R^s_k$ denotes the representative point of the cluster $R_k$, as:
\begin{equation}
    R^s_k= \frac{1}{\left\|R_k \right\|} \sum_{i \in R_k} F^s(I^s(i)).
\end{equation}
The $\left\|R_k \right\|$ represents the number of images belonging to cluster $R_k$. The centroids $R^s$ serve as essential reference points for comparing and selecting target images during the active sample selection process.

\vspace{3pt}\noindent\textbf{Active Sample Selection.} As depicted in Fig.~\ref{fig:STDR}, the model's parameters trained in the source domain are kept fixed and applied to process MR images in the target domain. The resulting features are then mapped to the latent space to obtain $F^t$, following the procedure described in Equation \ref{e3}. Next, we compute the Euclidean distances between $F^t(I^t(i))$ and all source-domain references $R^s$, and designate the smallest distance as the measure of the target-domain sample's similarity to the source domain:
\begin{equation}
    Similarity(I^t(i)) = \underset{k}{\min} {\left\|F^t(I^t(i)) - R^s_k \right\|}^2_2,   \label{eq6}
\end{equation}
where $F^t(\cdot)$ is similar to Equation \ref{e3} and represents the projection function for data in the target domain. 
In our source-free setting, the active learning sample selection encompasses both target domain representative samples $P^t$ and source domain representative samples $P^s$. This approach ensures the preservation of the sharing domain-invariant segmentation knowledge from the source domain, even during the migration to the target domain. the $Selected(P^s,P^t)$ is specified in the following way:
\begin{equation}
    Selected = Min(Similarity)[\frac{p}{2}\%]+Max(Similarity)[\frac{p}{2}\%],  \label{eq7}
\end{equation}
where the $similarity$ represents a list of the similarities of all the target domain data, and $p\%$ represents the proportion of samples to be selected for active learning. Our subsequent ablation experiments further demonstrate that this approach effectively preserves shared domain-invariant segmentation knowledge, leading to improved segmentation accuracy.

\subsection{Semi-supervised Learning for Source-free Active Domain Adaptation}

In our designed semi-supervised procedure, we employ a three-stage approach. The $\mathrm{stage}\ 1$ entails fully supervised fine-tuning of the model parameters using actively labeled samples $I^t_{AL}$. The detailed loss function is as follows:
\begin{equation}
L_s(\theta_t, \Omega_t) = \frac{1}{|\Omega_t|} \sum_{i=1}^{T} \Psi(y_t(i), p_t(i, \theta_t)),
\end{equation}
where the $\theta_t$ represent the fine-turning network parameters, $y_t(i) \in \{0, 1\}$ denotes the pixel labels from $Y^t_{AL}$ and $ p_t(i, \theta_t)$ is the predicted value of the network. The $\Psi$ is the same as Equation \ref{e2}. Subsequently, in $\mathrm{stage}\ 2$, we keep these fine-tuned model parameters fixed and utilize them to infer pseudo-labels $Y^t_{PL}$ for the unlabeled data $I^t_{UL}$. Finally, in $\mathrm{stage}\ 3$, we jointly fine-tune the model parameters using the inferred pseudo-labels $Y^t_{PL}$ and the actively labeled samples $Y^t_{AL}$. It should be noted that here we utilize the identical loss function as in $\mathrm{stage}\ 1$, with the exception that it incorporates more comprehensive data. 
\par This multi-stage process allows us to effectively leverage both labeled and unlabeled data. The initial fine-tuning of the actively labeled samples enables the model to learn the feature distribution of the target domain while preserving the shared domain-invariant segmentation knowledge. This establishes a strong foundation for the model, enabling it to generate higher-quality pseudo-labels. By using the fixed model to generate pseudo-labels $Y^t_{AL}$ for the unlabeled data $I^t_{UL}$, we expand the training dataset and gain additional valuable information. Finally, incorporating the inferred pseudo-labels alongside the actively selected samples in the last fine-tuning stage facilitates further refinement of the model, leading to improved performance and generalization. Our subsequent ablation experiments will further validate this aspect.

\subsection{Model Training and Implementation Details}
To maintain objectivity during evaluation, we partition each dataset randomly into three distinct subsets (train: valid: test) using a 7: 1: 2 ratio. In our setting, 70\% of the data from each medical center is allocated for model training and parameter tuning. Subsequently, we select the model with the best performance on the validation set and report its results on the test set. Considering the potential benefits of larger datasets in terms of richer segmentation samples and knowledge, we opt to designate the SMU center with the largest data volume, as our source domain. Subsequently, we utilized the datasets from the remaining four centers as target domains for Source-Free Active Domain Adaptation. Furthermore, to assess the generalization and robustness of our proposed method, we conduct additional experiments using data from the SPH center as the source domain for training. Subsequently, we migrate the trained model to other datasets using our approach to evaluate its performance and effectiveness.

Considering the huge different thicknesses of MR images from different hospitals, we segment GTV slice by slice in the axis slices and then stack them as volumetric predictions. We use 2D U-Net \cite{ronneberger2015u} as the baseline for all methods, the implementations are based on the PyMIC\cite{wang2023pymic}. All experiments are executed on an NVIDIA RTX 3090 GPU with 24 GB of memory. To standardize the training process, all model inputs are adjusted to a uniform resolution of 256$\times$256 and normalized the intensity to zero mean and unit variance. Furthermore, to boost the model's robustness, random rotation, flip and gaussian noise are applied for data augmentation. All models are trained using the SGD optimizer and a batch size of 32. When training with all data in the source or target domain, all models are trained with 30$k$ iterations. When training with actively selected samples, the data size is small and therefore 20$k$ iterations are trained. The initial learning rate is set to 0.03 and decays exponentially at a factor of 0.9 after each iteration. To be fair comparisons, we re-implement all comparison methods with the same backbone (U-Net) and run them in the same settings. For quantitative evaluation, we measure the volumetric-level Dice Similarity Coefficient, DSC ($\%$), 95\% Hausdorff distance, HD95 ($mm$), and Average Surface Distance, ASD ($mm$) between the ground truth and predictions. Generally, a superior method should yield a higher DSC
score and lower values for HD95 and ASD.

\section{Experiments and Results}
In this section, we first introduce the performance of the models trained in the respective data centers, and the performance of the trained models directly applied to other data centers. Subsequently, we compare our approach with state-of-the-art methods in a comprehensive manner, and finally we report ablation experimental results to demonstrate the effectiveness of our design. 
\subsection{Model training on source domain data}

\begin{table}[t]
\centering
\caption{Quantitative results of the model trained and tested on different datasets.}
\vspace{-1mm}
\label{tab2}
\resizebox{0.5\textwidth}{!}{%
\begin{tabular}{c|c|c c c }
\hline
Model & Dataset & DSC ($\%$) & HD95 ($mm$) & ASD ($mm$)  \\
\hline
U-Net & SCH & $72.19\pm 0.06$ & $9.25\pm 3.52$ & $2.22\pm 0.86$  \\
U-Net & APH & $86.79\pm 0.05$ & $4.92\pm 3.35$ & $1.13\pm 1.03$ \\
U-Net & SPH & $80.16\pm 0.08$ & $4.42\pm 1.63$ & $1.42\pm 0.71$ \\
U-Net & WCH & $85.87\pm 0.05$ & $4.15\pm 2.47$ & $1.17\pm 0.59$ \\
U-Net & SMU & $85.31\pm 0.07$ & $5.88\pm 14.83$ & $0.76\pm 0.69$ \\
\hline
\end{tabular}}
\vspace{-5mm}
\end{table}

\begin{table}[t]
\centering
\caption{Quantitative results  with models trained on different datasets and tested on other datasets.}
\vspace{-0.6mm}
\label{tab3}
\resizebox{0.5\textwidth}{!}{%
\begin{tabular}{c|c|c c c }
\hline
Model & Adaptation setting & DSC ($\%$) & HD95 ($mm$) & ASD ($mm$)  \\
\hline
U-Net & $\mathrm{SPH} \rightarrow \mathrm{SCH}$ & $59.11\pm 0.09$ & $31.35\pm 11.99$ & $3.13\pm 1.73$  \\
U-Net & $\mathrm{SPH} \rightarrow \mathrm{APH}$ & $78.25\pm 0.09$ & $9.14\pm 8.49$ & $2.68\pm 2.95$ \\
U-Net & $\mathrm{SPH} \rightarrow \mathrm{WCH}$ & $74.76\pm 0.11$ & $13.09\pm 12.78$ & $4.02\pm 4.44$ \\
U-Net & $\mathrm{SPH} \rightarrow \mathrm{SMU}$ & $73.71\pm 13.55$ & $12.89\pm 20.91$ & $3.41\pm 5.92$  \\
\hline
U-Net & $\mathrm{SMU} \rightarrow \mathrm{SCH}$ & $63.35\pm 0.10$ & $42.88\pm 17.29$ & $2.49\pm 2.69$ \\
U-Net & $\mathrm{SMU} \rightarrow \mathrm{APH}$ & $81.45\pm 8.12$ & $8.31\pm 8.97$ & $2.60\pm 3.37$ \\
U-Net & $\mathrm{SMU} \rightarrow \mathrm{SPH}$ & $77.51\pm 8.49$ & $9.42\pm 17.59$ & $2.96\pm 4.95$ \\
U-Net & $\mathrm{SMU} \rightarrow \mathrm{WCH}$ & $77.40\pm 8.47$ & $7.94\pm 7.07$ & $2.37\pm 1.70$ \\
\hline
\end{tabular}}
\vspace{-5mm}
\end{table}

Initially, we train the models on the training sets of the five medical centers separately. The selected models from the validation sets are subsequently evaluated on the corresponding test sets of each medical center. The results of these evaluations are presented in Table~\ref{tab2}. The results indicate that even the same model and training strategy could produce varying outcomes across different data centers. Specifically, the segmentation result at SCH achieved a DSC of only 72.19\%, which may be attributed to more intricate patient conditions and smaller dataset sizes. In contrast, the other four centers yielded a DSC of over 80\%. In addition, our findings suggest that a larger dataset does not necessarily yield superior results. For instance, the performance at APH, which only had 146 patients, outperforms that at SPH, which had 208 patients.

Table~\ref{tab3} presents the performances of deploying the segmentation model to other medical centers directly, after training it on SPH and SMU datasets. Our results demonstrate that the existence of domain gaps leads to a decrease in DSC accuracy of 5-10\% in the majority of cases, such as $\mathrm{SPH} \rightarrow \mathrm{APH}$ and $\mathrm{SMU} \rightarrow \mathrm{SCH}$. In some cases, however, we observed a DSC accuracy degradation of more than 10\%, such as $\mathrm{SPH} \rightarrow \mathrm{SCH}$ and $\mathrm{SPH} \rightarrow \mathrm{SMU}$. This further confirms that deploying computer-aided GTV segmentation algorithms in different medical centers requires active source-free domain adaptation of the model. In addition, we can also find that the model transferred from SMU performed better than the one from SPH. This observation inspires us that in practical applications, we can adapt the model trained on a dataset with a larger volume of data to smaller data centers using source-free active domain adaptation to yield superior results.

\begin{table}[t]
\centering
\caption{Quantitative results  with models trained on SPH dataset and adapted on other datasets.}
\vspace{-0.6mm}
\label{tab4}
\resizebox{0.5\textwidth}{!}{%
\begin{tabular}{c|c|c| c c c }
\hline
Model & Adaptation setting & Source data &DSC ($\%$) & HD95 ($mm$) & ASD ($mm$)  \\
\hline   
AdaptSeg \cite{tsai2018learning} & $\mathrm{SPH} \rightarrow \mathrm{SCH}$ & Yes & $59.63\pm 11.03$ & $27.84\pm 17.52$ & $7.31\pm 5.12$  \\
AdaptSeg \cite{tsai2018learning} & $\mathrm{SPH} \rightarrow \mathrm{APH}$ & Yes & $77.87\pm 15.26$ & $7.62\pm 7.19$ & $1.97\pm 1.43$ \\
AdaptSeg \cite{tsai2018learning} & $\mathrm{SPH} \rightarrow \mathrm{WCH}$ & Yes & $74.95\pm 8.64$ & $10.74\pm 7.71$ & $3.64\pm 2.45$ \\
AdaptSeg \cite{tsai2018learning} & $\mathrm{SPH} \rightarrow \mathrm{SMU}$ & Yes & $74.38\pm 12.58$ & $12.05\pm 21.64$ & $3.11\pm 4.97$  \\
\hline   
AdvEnt \cite{vu2019advent} & $\mathrm{SPH} \rightarrow \mathrm{SCH}$ & Yes & $60.14\pm 10.16$ & $20.49\pm 16.21$ & $4.65\pm 2.72$ \\
AdvEnt \cite{vu2019advent} & $\mathrm{SPH} \rightarrow \mathrm{APH}$ & Yes & $79.71\pm 11.37$ & $7.13\pm 5.67$ & $2.18\pm 1.63$ \\
AdvEnt \cite{vu2019advent} & $\mathrm{SPH} \rightarrow \mathrm{WCH}$ & Yes & $75.27\pm 8.22$ & $9.94\pm 6.69$ & $3.19\pm 2.08$ \\
AdvEnt \cite{vu2019advent} & $\mathrm{SPH} \rightarrow \mathrm{SMU}$ & Yes & $75.56\pm 11.47$ & $10.41\pm 19.32$ & $2.79\pm 4.68$ \\
\hline   
UncertainDA \cite{fleuret2021uncertainty} & $\mathrm{SPH} \rightarrow \mathrm{SCH}$ & No & $61.74\pm 11.24$ & $21.29\pm 16.91$ & $4.84\pm 2.72$  \\
UncertainDA \cite{fleuret2021uncertainty} & $\mathrm{SPH} \rightarrow \mathrm{APH}$ & No & $80.03\pm 11.56$ & $6.81\pm 5.22$ & $2.01\pm 1.76$ \\
UncertainDA \cite{fleuret2021uncertainty} & $\mathrm{SPH} \rightarrow \mathrm{WCH}$ & No & $75.88\pm 8.81$ & $12.14\pm 12.50$ & $3.65\pm 4.02$ \\
UncertainDA \cite{fleuret2021uncertainty} & $\mathrm{SPH} \rightarrow \mathrm{SMU}$ & No & $76.02\pm 12.67$ & $9.45\pm 18.96$ & $2.06\pm 4.57$  \\
\hline     
Tent \cite{wang2020tent} & $\mathrm{SPH} \rightarrow \mathrm{SCH}$ & No & $62.37\pm 9.74$ & $26.02\pm 17.27$ & $5.45\pm 5.37$ \\
Tent \cite{wang2020tent} & $\mathrm{SPH} \rightarrow \mathrm{APH}$ & No & $80.73\pm 10.46$ & $6.85\pm 6.15$ & $1.99\pm 1.81$ \\
Tent \cite{wang2020tent} & $\mathrm{SPH} \rightarrow \mathrm{WCH}$ & No & $75.81\pm 10.31$ & $11.24\pm 10.51$ & $3.51\pm 3.57$  \\
Tent \cite{wang2020tent} & $\mathrm{SPH} \rightarrow \mathrm{SMU}$ & No & $75.88\pm 12.15$ & $8.17\pm 17.52$ & $1.94\pm 3.14$ \\
\hline    
DPL \cite{chen2021source} & $\mathrm{SPH} \rightarrow \mathrm{SCH}$ & No & $62.88\pm 10.36$ & $21.29\pm 14.56$ & $4.27\pm 3.16$  \\
DPL \cite{chen2021source} & $\mathrm{SPH} \rightarrow \mathrm{APH}$ & No & $80.96\pm 12.14$ & $6.23\pm 5.15$ & $1.78\pm 1.49$ \\
DPL \cite{chen2021source} & $\mathrm{SPH} \rightarrow \mathrm{WCH}$ & No & $77.26\pm 9.83$ & $8.96\pm 5.73$ & $2.81\pm 1.55$ \\
DPL \cite{chen2021source} & $\mathrm{SPH} \rightarrow \mathrm{SMU}$ & No & $77.24\pm 10.53$ & $9.22\pm 18.77$ & $1.85\pm 3.23$  \\
\hline
Ours & $\mathrm{SPH} \rightarrow \mathrm{SCH}$ & No & $\textbf{70.99}\pm 6.59$ & $\textbf{8.57}\pm 2.39$ & $\textbf{2.03}\pm 0.84$ \\
Ours & $\mathrm{SPH} \rightarrow \mathrm{APH}$ & No & $\textbf{85.69}\pm 4.54$ & $\textbf{4.58}\pm 2.24$ & $\textbf{1.18}\pm 0.78$ \\
Ours & $\mathrm{SPH} \rightarrow \mathrm{WCH}$ & No & $\textbf{83.59}\pm 6.76$ & $\textbf{4.85}\pm 2.79$ & $\textbf{1.37}\pm 0.92$ \\
Ours & $\mathrm{SPH} \rightarrow \mathrm{SMU}$ & No & $\textbf{84.13}\pm 7.65$ & $\textbf{6.17}\pm 15.33$ & $\textbf{0.86}\pm 0.95$ \\
\hline
\end{tabular}}
\vspace{-5mm}
\end{table}

\subsection{Comparison with State-of-the-art Methods}
To ensure a thorough and comprehensive evaluation of our method's performance, we extended our analysis beyond merely comparing it with other SOTA domain adaptation methods (Tables~\ref{tab4} and \ref{tab5}). We also conducted a performance comparison with SOTA active learning methods (Tables~\ref{tab6} and \ref{tab7}). Under both comparisons, we conducted extensive experiments containing two kinds of data settings ($\mathrm{SPH} \rightarrow$ other datasets, $\mathrm{SMU} \rightarrow$ other datasets). Our methods consistently demonstrated superior performance across all experiments, affirming the effectiveness of our approach.

\subsubsection{Comparison with State-of-the-art Domain adaptation methods}
\begin{figure*}[t]
    \centering
    \includegraphics[width=0.96\textwidth]{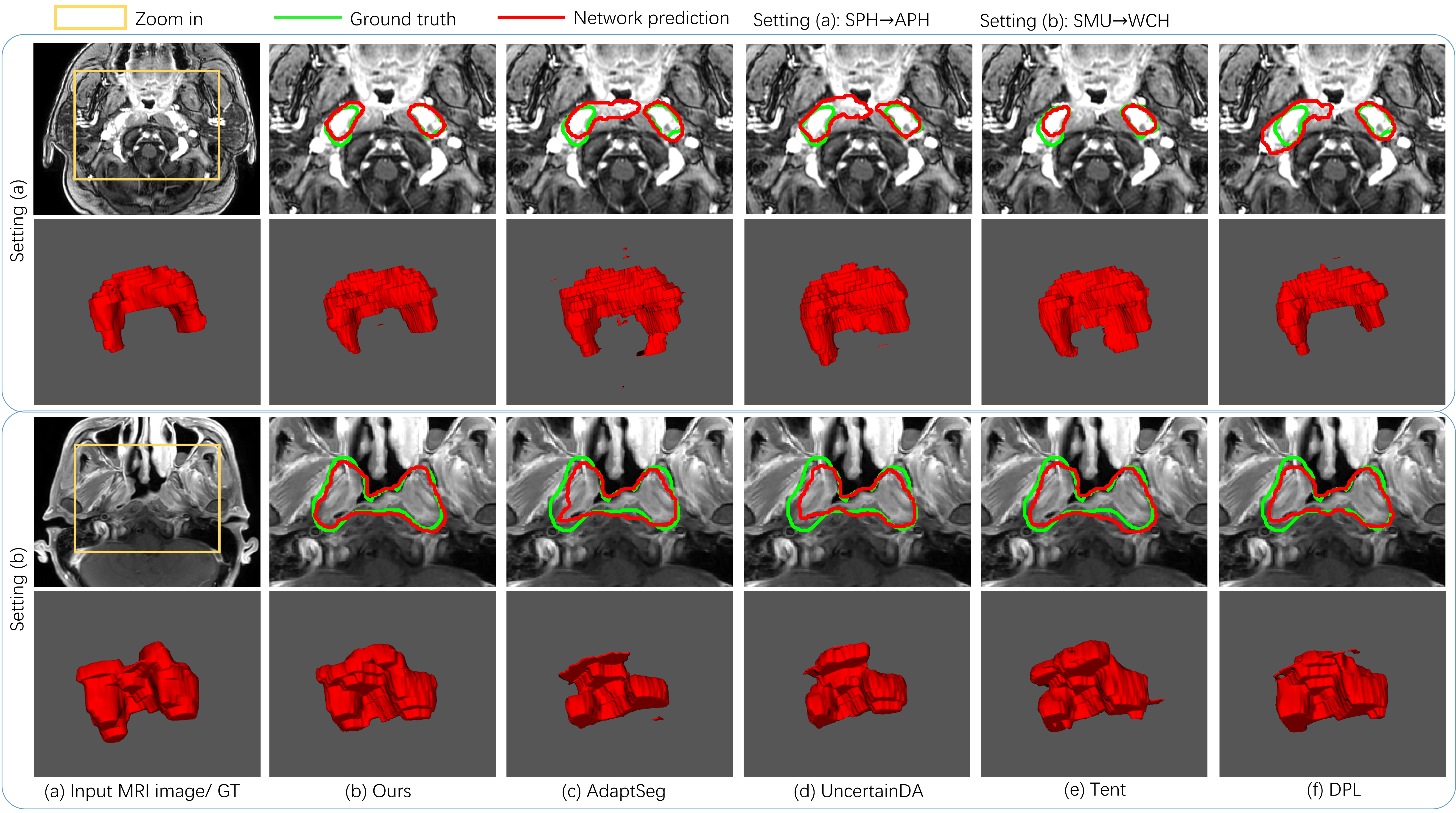}
    \vspace{-1mm}
    \caption{Visual comparisons of (b) our method with other SOTA methods. The first and third rows showcase 2D slice visualizations, while the second and fourth rows offer 3D comparisons. Contours exhibit markedly improved alignment with ground truth both at the 2D and 3D levels following the application of our method, as compared to other methods.}
    \label{fig:demo}
    \vspace{-3mm}
\end{figure*}

We conduct a comparative analysis between our approach and SOTA domain adaptation methods to assess their performance. This evaluation encompasses methods with (AdaptSeg \cite{tsai2018learning}, AdvEnt \cite{vu2019advent}) and without (UncertainDA \cite{fleuret2021uncertainty}, Tent \cite{wang2020tent} and DPL \cite{chen2021source}) access to source domain data. As anticipated, our proposed method exhibits significant enhancements compared to the other domain adaptation methods in both Table~\ref{tab4} and Table~\ref{tab5}. This outcome underscores the impactful role of strategically selected active samples, in tandem with the semi-supervised pipeline, in achieving substantial performance advancements. As an illustration, our method demonstrates a superiority over the best DSC achieved by other approaches, surpassing them by 8.11\% and 8.07\% points on $\mathrm{SPH}\rightarrow\mathrm{SCH}$ and $\mathrm{SMU}\rightarrow\mathrm{SCH}$, respectively. Fig.~\ref{fig:demo} additionally provides an intuitive visual representation of segmentation instances using various methods.

\begin{table}[t]
\centering
\caption{Quantitative results  with models trained on SMU dataset and adapted on other datasets.}
\vspace{-0.6mm}
\label{tab5}
\resizebox{0.5\textwidth}{!}{%
\begin{tabular}{c|c|c| c c c }
\hline
Model & Adaptation setting & Source data &DSC ($\%$) & HD95 ($mm$) & ASD ($mm$)  \\
\hline
AdaptSeg \cite{tsai2018learning} & $\mathrm{SMU} \rightarrow \mathrm{SCH}$ & Yes & $63.56\pm 10.57$ & $41.38\pm 15.13$ & $2.65\pm 2.73$  \\
AdaptSeg \cite{tsai2018learning} & $\mathrm{SMU} \rightarrow \mathrm{APH}$ & Yes & $82.11\pm 7.15$ & $8.26\pm 8.67$ & $2.80\pm 2.21$ \\
AdaptSeg \cite{tsai2018learning} & $\mathrm{SMU} \rightarrow \mathrm{SPH}$ & Yes & $76.87\pm 9.62$ & $8.85\pm 11.99$ & $2.57\pm 2.07$ \\
AdaptSeg \cite{tsai2018learning} & $\mathrm{SMU} \rightarrow \mathrm{WCH}$ & Yes & $77.95\pm 8.57$ & $7.39\pm 4.86$ & $2.00\pm 1.19$  \\
\hline
AdvEnt \cite{vu2019advent} & $\mathrm{SMU} \rightarrow \mathrm{SCH}$ & Yes & $64.06\pm 7.57$ & $36.18\pm 14.12$ & $2.90\pm 1.76$ \\
AdvEnt \cite{vu2019advent} & $\mathrm{SMU} \rightarrow \mathrm{APH}$ & Yes & $80.91\pm 6.31$ & $6.91\pm 5.93$ & $2.33\pm 1.97$ \\
AdvEnt \cite{vu2019advent} & $\mathrm{SMU} \rightarrow \mathrm{SPH}$ & Yes & $76.71\pm 8.99$ & $6.97\pm 11.29$ & $1.90\pm 1.75$ \\
AdvEnt \cite{vu2019advent} & $\mathrm{SMU} \rightarrow \mathrm{WCH}$ & Yes & $78.37\pm 9.03$ & $7.15\pm 4.84$ & $1.95\pm 1.48$ \\
\hline
UncertainDA \cite{fleuret2021uncertainty} & $\mathrm{SMU} \rightarrow \mathrm{SCH}$ & No & $63.59\pm 8.32$ & $28.73\pm 12.28$ & $4.38\pm 2.85$  \\
UncertainDA \cite{fleuret2021uncertainty} & $\mathrm{SMU} \rightarrow \mathrm{APH}$ & No & $81.17\pm 7.11$ & $7.96\pm 4.05$ & $2.24\pm 1.28$ \\
UncertainDA \cite{fleuret2021uncertainty} & $\mathrm{SMU} \rightarrow \mathrm{SPH}$ & No & $77.88\pm 9.58$ & $6.44\pm 11.56$ & $1.89\pm 1.79$ \\
UncertainDA \cite{fleuret2021uncertainty} & $\mathrm{SMU} \rightarrow \mathrm{WCH}$ & No & $78.16\pm 9.10$ & $6.67\pm 4.44$ & $1.87\pm 1.07$  \\
\hline
Tent \cite{wang2020tent} & $\mathrm{SMU} \rightarrow \mathrm{SCH}$ & No & $64.43\pm 7.55$ & $29.25\pm 16.34$ & $3.39\pm 2.23$ \\
Tent \cite{wang2020tent} & $\mathrm{SMU} \rightarrow \mathrm{APH}$ & No & $80.49\pm 7.58$ & $7.94\pm 7.14$ & $2.48\pm 2.56$ \\
Tent \cite{wang2020tent} & $\mathrm{SMU} \rightarrow \mathrm{SPH}$ & No & $78.46\pm 9.15$ & $7.37\pm 12.02$ & $2.18\pm 2.44$ \\
Tent \cite{wang2020tent} & $\mathrm{SMU} \rightarrow \mathrm{WCH}$ & No & $78.13\pm 8.79$ & $6.46\pm 4.09$ & $1.65\pm 0.99$ \\
\hline
DPL \cite{chen2021source} & $\mathrm{SMU} \rightarrow \mathrm{SCH}$ & No & $64.51\pm 7.70$ & $24.74\pm 7.27$ & $3.83\pm 1.85$  \\
DPL \cite{chen2021source} & $\mathrm{SMU} \rightarrow \mathrm{APH}$ & No & $81.43\pm 6.74$ & $6.38\pm 5.68$ & $1.97\pm 2.02$ \\
DPL \cite{chen2021source} & $\mathrm{SMU} \rightarrow \mathrm{SPH}$ & No & $78.83\pm 8.38$ & $6.89\pm 12.67$ & $1.94\pm 2.95$ \\
DPL \cite{chen2021source} & $\mathrm{SMU} \rightarrow \mathrm{WCH}$ & No & $78.92\pm 8.42$ & $6.26\pm 3.79$ & $1.72\pm 1.04$  \\
\hline
Ours & $\mathrm{SMU} \rightarrow \mathrm{SCH}$ & No & $\textbf{72.58}\pm 6.72$ & $\textbf{8.06}\pm 2.06$ & $\textbf{2.00}\pm 0.89$ \\
Ours & $\mathrm{SMU} \rightarrow \mathrm{APH}$ & No & $\textbf{86.01}\pm 5.10$ & $\textbf{4.38}\pm 2.51$ & $\textbf{1.21}\pm 1.04$ \\
Ours & $\mathrm{SMU} \rightarrow \mathrm{SPH}$ & No & $\textbf{80.97}\pm 7.24$ & $\textbf{4.02}\pm 1.62$ & $\textbf{1.03}\pm 0.58$ \\
Ours & $\mathrm{SMU} \rightarrow \mathrm{WCH}$ & No & $\textbf{84.71}\pm 6.04$ & $\textbf{4.48}\pm 2.23$ & $\textbf{1.24}\pm 0.77$ \\
\hline
\end{tabular}}
\vspace{-5mm}
\end{table}

\begin{table}[t]
\centering
\caption{Experimental results  on various active sample selection methods from SPH dataset to other datasets.}
\vspace{-0.6mm}
\label{tab6}
\resizebox{0.5\textwidth}{!}{%
\begin{tabular}{c|c| c c c }
\hline
Model & Adaptation setting &DSC ($\%$) & HD95 ($mm$) & ASD ($mm$)  \\
\hline
Random & $\mathrm{SPH} \rightarrow \mathrm{SCH}$  & $62.28\pm 9.72$ & $26.48\pm 10.59$ & $2.64\pm 1.61$  \\
Random & $\mathrm{SPH} \rightarrow \mathrm{APH}$  & $81.19\pm 8.69$ & $6.55\pm 4.82$ & $2.01\pm 1.5$ \\
Random & $\mathrm{SPH} \rightarrow \mathrm{WCH}$  & $78.05\pm 7.40$ & $6.53\pm 6.20$ & $1.92\pm 1.94$ \\
Random & $\mathrm{SPH} \rightarrow \mathrm{SMU}$  & $78.75\pm 10.26$ & $7.36\pm 14.77$ & $1.37\pm 2.45$  \\
\hline    
Adversarial \cite{tsai2018learning} & $\mathrm{SPH} \rightarrow \mathrm{SCH}$ & $63.21\pm 7.23$ & $24.75\pm 8.15$ & $3.42\pm 1.61$ \\
Adversarial \cite{tsai2018learning} & $\mathrm{SPH} \rightarrow \mathrm{APH}$ & $82.21\pm 6.32$ & $8.27\pm 7.43$ & $2.35\pm 1.81$ \\
Adversarial \cite{tsai2018learning} & $\mathrm{SPH} \rightarrow \mathrm{WCH}$ & $79.00\pm 7.87$ & $7.07\pm 5.20$ & $2.21\pm 1.42$ \\
Adversarial \cite{tsai2018learning} & $\mathrm{SPH} \rightarrow \mathrm{SMU}$ & $79.44\pm 9.39$ & $7.17\pm 12.57$ & $1.58\pm 1.51$ \\
\hline    
Entropy \cite{vu2019advent} & $\mathrm{SPH} \rightarrow \mathrm{SCH}$ & $65.02\pm 8.67$ & $18.03\pm 7.34$ & $3.34\pm 1.87$  \\
Entropy \cite{vu2019advent} & $\mathrm{SPH} \rightarrow \mathrm{APH}$ & $83.45\pm 5.59$ & $7.59\pm 9.45$ & $1.82\pm 2.12$ \\
Entropy \cite{vu2019advent} & $\mathrm{SPH} \rightarrow \mathrm{WCH}$ & $80.97\pm 7.98$ & $7.50\pm 5.98$ & $2.22\pm 1.37$ \\
Entropy \cite{vu2019advent} & $\mathrm{SPH} \rightarrow \mathrm{SMU}$ & $80.27\pm 10.34$ & $7.05\pm 14.74$ & $1.43\pm 2.02$  \\
\hline    
AADA \cite{su2020active} & $\mathrm{SPH} \rightarrow \mathrm{SCH}$ & $65.08\pm 9.82$ & $14.45\pm 5.90$ & $2.88\pm 1.67$ \\
AADA \cite{su2020active} & $\mathrm{SPH} \rightarrow \mathrm{APH}$ & $84.14\pm 5.05$ & $7.87\pm 7.66$ & $1.91\pm 1.95$ \\
AADA \cite{su2020active} & $\mathrm{SPH} \rightarrow \mathrm{WCH}$ & $81.21\pm 7.23$ & $6.52\pm 3.81$ & $2.01\pm 1.05$ \\
AADA \cite{su2020active} & $\mathrm{SPH} \rightarrow \mathrm{SMU}$ & $80.78\pm 8.60$ & $6.33\pm 12.53$ & $\textbf{1.12}\pm 1.62$ \\
\hline
STDR & $\mathrm{SPH} \rightarrow \mathrm{SCH}$ & $\textbf{69.55}\pm 0.07$ & $\textbf{10.64}\pm 3.36$ & $\textbf{2.83}\pm 1.83$  \\
STDR & $\mathrm{SPH} \rightarrow \mathrm{APH}$ & $\textbf{85.05}\pm 0.05$ & $\textbf{4.98}\pm 3.07$ & $\textbf{1.36}\pm 1.01$ \\
STDR & $\mathrm{SPH} \rightarrow \mathrm{WCH}$ & $\textbf{82.81}\pm 0.07$ & $\textbf{5.39}\pm 2.77$ & $\textbf{1.58}\pm 0.88$ \\
STDR & $\mathrm{SPH} \rightarrow \mathrm{SMU}$ & $\textbf{83.51}\pm 0.08$ & $\textbf{5.81}\pm 13.10$ & $1.26\pm 1.56$  \\
\hline
\end{tabular}}
\vspace{-5mm}
\end{table}

\begin{table}[t]
\centering
\caption{Experimental results  on various active sample selection methods from SMU dataset to other datasets.}
\vspace{-0.6mm}
\label{tab7}
\resizebox{0.5\textwidth}{!}{%
\begin{tabular}{c|c| c c c }
\hline
Model & Adaptation setting &DSC ($\%$) & HD95 ($mm$) & ASD ($mm$)  \\
\hline
Random & $\mathrm{SMU} \rightarrow \mathrm{SCH}$ & $66.67\pm 7.51$ & $15.06\pm 7.28$ & $3.35\pm 2.23$  \\
Random & $\mathrm{SMU} \rightarrow \mathrm{APH}$ & $82.66\pm 6.37$ & $7.28\pm 9.62$ & $2.26\pm 1.79$ \\
Random & $\mathrm{SMU} \rightarrow \mathrm{SPH}$ & $78.89\pm 8.79$ & $6.66\pm 11.57$ & $1.81\pm 1.88$ \\
Random & $\mathrm{SMU} \rightarrow \mathrm{WCH}$ & $79.77\pm 8.01$ & $7.58\pm 3.95$ & $2.38\pm 1.15$  \\
\hline 
Adversarial \cite{tsai2018learning} & $\mathrm{SMU} \rightarrow \mathrm{SCH}$ & $67.67\pm 7.20$ & $19.85\pm 9.63$ & $3.12\pm 2.31$ \\
Adversarial \cite{tsai2018learning} & $\mathrm{SMU} \rightarrow \mathrm{APH}$ & $82.47\pm 7.23$ & $7.16\pm 6.96$ & $2.03\pm 1.94$ \\
Adversarial \cite{tsai2018learning} & $\mathrm{SMU} \rightarrow \mathrm{SPH}$ & $79.07\pm 7.79$ & $9.08\pm 12.71$ & $2.42\pm 2.51$ \\
Adversarial \cite{tsai2018learning} & $\mathrm{SMU} \rightarrow \mathrm{WCH}$ & $81.18\pm 8.29$ & $8.83\pm 10.41$ & $2.83\pm 3.56$ \\
\hline
Entropy \cite{vu2019advent} & $\mathrm{SMU} \rightarrow \mathrm{SCH}$ & $67.78\pm 6.88$ & $17.68\pm 8.55$ & $4.77\pm 2.03$  \\
Entropy \cite{vu2019advent} & $\mathrm{SMU} \rightarrow \mathrm{APH}$ & $83.24\pm 7.01$ & $7.41\pm 8.46$ & $2.15\pm 2.80$ \\
Entropy \cite{vu2019advent} & $\mathrm{SMU} \rightarrow \mathrm{SPH}$ & $79.11\pm 8.82$ & $6.99\pm 11.75$ & $1.85\pm 1.91$ \\
Entropy \cite{vu2019advent} & $\mathrm{SMU} \rightarrow \mathrm{WCH}$ & $81.88\pm 7.38$ & $6.03\pm 6.68$ & $1.99\pm 2.74$  \\
\hline
AADA \cite{su2020active} & $\mathrm{SMU} \rightarrow \mathrm{SCH}$ & $67.96\pm 5.15$ & $13.86\pm 9.73$ & $2.53\pm 1.67$ \\
AADA \cite{su2020active} & $\mathrm{SMU} \rightarrow \mathrm{APH}$ & $83.21\pm 5.83$ & $6.31\pm 7.03$ & $1.83\pm 2.18$ \\
AADA \cite{su2020active} & $\mathrm{SMU} \rightarrow \mathrm{SPH}$ & $79.77\pm 7.78$ & $6.50\pm 5.79$ & $1.77\pm 1.36$ \\
AADA \cite{su2020active} & $\mathrm{SMU} \rightarrow \mathrm{WCH}$ & $81.73\pm 8.35$ & $5.70\pm 3.63$ & $1.74\pm 1.44$ \\
\hline
STDR & $\mathrm{SMU} \rightarrow \mathrm{SCH}$ & $\textbf{70.99}\pm 6.46$ & $\textbf{10.72}\pm 3.37$ & $\textbf{2.47}\pm 1.07$  \\
STDR & $\mathrm{SMU} \rightarrow \mathrm{APH}$ & $\textbf{85.49}\pm 6.09$ & $\textbf{4.63}\pm 3.06$ & $\textbf{1.24}\pm 1.02$ \\
STDR & $\mathrm{SMU} \rightarrow \mathrm{SPH}$ & $\textbf{80.00}\pm 8.10$ & $\textbf{4.49}\pm 2.64$ & $\textbf{1.64}\pm 1.46$ \\
STDR & $\mathrm{SMU} \rightarrow \mathrm{WCH}$ & $\textbf{84.11}\pm 6.64$ & $\textbf{4.95}\pm 2.75$ & $\textbf{1.35}\pm 0.88$  \\
\hline
\end{tabular}}
\vspace{-5mm}
\end{table}

\subsubsection{Comparison with State-of-the-art Active learning methods}
Meanwhile, considering that our active domain adaptation framework inherently incorporates the concept of active learning, it is reasonable to conduct a comparison with other SOTA active learning methods. Therefore, we opt to conduct a comparison between our approach and random selection, alongside three SOTA active learning methods. The details are as follows: (i) Random Selection: Samples are randomly chosen from the target domain with uniform probability; (ii) Adversarial \cite{tsai2018learning}: Leveraging the trained discriminator following \cite{tsai2018learning}, target samples are selected with the lowest predicted probabilities, indicating those that exhibit the most pronounced divergence from the source domain; (iii) Entropy \cite{vu2019advent}: The AdvEnt method \cite{vu2019advent} is employed to calculate the prediction map entropy for each sample within the target domain, and those samples with the highest entropy are selected for manual annotation; (iv) AADA \cite{su2020active}: AADA considers both sample uncertainty and diversity when making selections.

For equitable comparisons, the identical experimental framework is utilized across all comparative experiments. Each experiment involves selecting only 20\% active samples for labeling and training. These experimental results in Table~\ref{tab6} and Table~\ref{tab7} demonstrate that the devised STDR strategy provides the best segmentation performance. This proves the superiority of the proposed strategy and the effectiveness of considering domain-invariant and domain-specific representations in sample selection.

\begin{table}[t]
\centering
\caption{Results  of ablation experiments from SPH dataset to other datasets.}
\vspace{-0.6mm}
\label{tab8}
\resizebox{0.5\textwidth}{!}{%
\begin{tabular}{c|c| c c c }
\hline
Model & Adaptation setting &DSC ($\%$) & HD95 ($mm$) & ASD ($mm$)  \\
\hline
STDR-$\alpha$ & $\mathrm{SPH} \rightarrow \mathrm{SCH}$ & $62.66\pm 11.33$ & $16.48\pm 16.03$ & $4.54\pm 4.02$  \\
STDR-$\alpha$ & $\mathrm{SPH} \rightarrow \mathrm{APH}$ & $81.46\pm 8.97$ & $5.64\pm .344$ & $1.57\pm 1.19$ \\
STDR-$\alpha$ & $\mathrm{SPH} \rightarrow \mathrm{WCH}$ & $79.51\pm 9.11$ & $5.63\pm 3.23$ & $1.58\pm 0.93$ \\
STDR-$\alpha$ & $\mathrm{SPH} \rightarrow \mathrm{SMU}$ & $77.03\pm 13.11$ & $8.45\pm 16.58$ & $1.23\pm 1.34$  \\
\hline
STDR-$\beta$ & $\mathrm{SPH} \rightarrow \mathrm{SCH}$ & $69.28\pm 7.78$ & $14.93\pm 9.12$ & $3.43\pm 2.71$ \\
STDR-$\beta$ & $\mathrm{SPH} \rightarrow \mathrm{APH}$ & $84.38\pm 5.41$ & $6.52\pm 7.48$ & $1.89\pm 1.88$ \\
STDR-$\beta$ & $\mathrm{SPH} \rightarrow \mathrm{WCH}$ & $81.79\pm 7.31$ & $6.06\pm 3.89$ & $1.77\pm 0.97$ \\
STDR-$\beta$ & $\mathrm{SPH} \rightarrow \mathrm{SMU}$ & $82.66\pm 9.14$ & $6.69\pm 15.83$ & $1.17\pm 1.41$ \\
\hline
STDR & $\mathrm{SPH} \rightarrow \mathrm{SCH}$ & $69.55\pm 0.07$ & $10.64\pm 3.36$ & $2.83\pm 1.83$  \\
STDR & $\mathrm{SPH} \rightarrow \mathrm{APH}$ & $85.05\pm 0.05$ & $4.98\pm 3.07$ & $1.36\pm 1.01$ \\
STDR & $\mathrm{SPH} \rightarrow \mathrm{WCH}$ & $82.81\pm 0.07$ & $5.39\pm 2.77$ & $1.58\pm 0.88$ \\
STDR & $\mathrm{SPH} \rightarrow \mathrm{SMU}$ & $83.51\pm 0.08$ & $\textbf{5.81}\pm 13.10$ & $1.26\pm 1.56$  \\
\hline
Ours (STDR+Semi) & $\mathrm{SPH} \rightarrow \mathrm{SCH}$ & $\textbf{70.99}\pm 6.59$ & $\textbf{8.57}\pm 2.39$ & $\textbf{2.03}\pm 0.84$ \\
Ours (STDR+Semi) & $\mathrm{SPH} \rightarrow \mathrm{APH}$ & $\textbf{85.69}\pm 4.54$ & $\textbf{4.58}\pm 2.24$ & $\textbf{1.18}\pm 0.78$ \\
Ours (STDR+Semi) & $\mathrm{SPH} \rightarrow \mathrm{WCH}$ & $\textbf{83.59}\pm 6.76$ & $\textbf{4.85}\pm 2.79$ & $\textbf{1.37}\pm 0.92$ \\
Ours (STDR+Semi) & $\mathrm{SPH} \rightarrow \mathrm{SMU}$ & $\textbf{84.13}\pm 7.65$ & $6.17\pm 15.33$ & $\textbf{0.86}\pm 0.95$ \\
\hline
\end{tabular}}
\vspace{-5mm}
\end{table}

\begin{table}[t]
\centering
\caption{Results  of ablation experiments from SMU dataset to other datasets.}
\vspace{-0.6mm}
\label{tab9}
\resizebox{0.5\textwidth}{!}{%
\begin{tabular}{c|c| c c c }
\hline
Model & Adaptation setting &DSC ($\%$) & HD95 ($mm$) & ASD ($mm$)  \\
\hline
STDR-$\alpha$ & $\mathrm{SMU} \rightarrow \mathrm{SCH}$  & $65.64\pm 7.84$ & $16.02\pm 9.75$ & $2.98\pm 1.55$  \\
STDR-$\alpha$ & $\mathrm{SMU} \rightarrow \mathrm{APH}$  & $81.57\pm 7.89$ & $6.78\pm 4.65$ & $1.94\pm 1.16$ \\
STDR-$\alpha$ & $\mathrm{SMU} \rightarrow \mathrm{SPH}$  & $76.80\pm 8.51$ & $5.29\pm 2.56$ & $1.67\pm 0.99$ \\
STDR-$\alpha$ & $\mathrm{SMU} \rightarrow \mathrm{WCH}$  & $79.44\pm 8.27$ & $6.48\pm 4.97$ & $1.68\pm 1.12$  \\
\hline
STDR-$\beta$ & $\mathrm{SMU} \rightarrow \mathrm{SCH}$  & $68.21\pm 7.16$ & $10.58\pm 3.35$ & $2.99\pm 1.67$ \\
STDR-$\beta$ & $\mathrm{SMU} \rightarrow \mathrm{APH}$  & $83.57\pm 6.57$ & $6.17\pm 8.25$ & $1.93\pm 2.63$ \\
STDR-$\beta$ & $\mathrm{SMU} \rightarrow \mathrm{SPH}$  & $79.86\pm 7.97$ & $5.42\pm 4.92$ & $1.48\pm 1.21$ \\
STDR-$\beta$ & $\mathrm{SMU} \rightarrow \mathrm{WCH}$  & $83.55\pm 6.69$ & $6.53\pm 5.33$ & $1.70\pm 1.14$ \\
\hline
STDR & $\mathrm{SMU} \rightarrow \mathrm{SCH}$  & $70.99\pm 6.46$ & $10.72\pm 3.37$ & $2.47\pm 1.07$  \\
STDR & $\mathrm{SMU} \rightarrow \mathrm{APH}$  & $85.49\pm 6.09$ & $4.63\pm 3.06$ & $1.24\pm 1.02$ \\
STDR & $\mathrm{SMU} \rightarrow \mathrm{SPH}$  & $80.00\pm 8.10$ & $4.49\pm 2.64$ & $1.64\pm 1.46$ \\
STDR & $\mathrm{SMU} \rightarrow \mathrm{WCH}$  & $84.11\pm 6.64$ & $4.95\pm 2.75$ & $1.35\pm 0.88$  \\
\hline
Ours (STDR+Semi) & $\mathrm{SMU} \rightarrow \mathrm{SCH}$ & $\textbf{72.58}\pm 6.72$ & $\textbf{8.06}\pm 2.06$ & $\textbf{2.00}\pm 0.89$ \\
Ours (STDR+Semi) & $\mathrm{SMU} \rightarrow \mathrm{APH}$ & $\textbf{86.01}\pm 5.10$ & $\textbf{4.38}\pm 2.51$ & $\textbf{1.21}\pm 1.04$ \\
Ours (STDR+Semi) & $\mathrm{SMU} \rightarrow \mathrm{SPH}$ & $\textbf{80.97}\pm 7.24$ & $\textbf{4.02}\pm 1.62$ & $\textbf{1.03}\pm 0.58$ \\
Ours (STDR+Semi) & $\mathrm{SMU} \rightarrow \mathrm{WCH}$ & $\textbf{84.71}\pm 6.04$ & $\textbf{4.48}\pm 2.23$ & $\textbf{1.24}\pm 0.77$ \\
\hline
\end{tabular}}
\vspace{-5mm}
\end{table}

\subsection{Ablation Studies}
To ascertain the effectiveness of the key components within our proposed methodology, we perform ablation experiments, leading to the identification of four distinct configurations: (i) STDR-$\alpha$: we select the 20\% samples with the smallest $Similarity$ in Eq.~\ref{eq6} and focus on domain-invariant representations; (ii) STDR-$\beta$: we choose the 20\% samples with the largest $Similarity$ in Eq.~\ref{eq6} and focus on domain-specific representations; (iii) STDR: we follow Eq.~\ref{eq7} to take both into consideration; (iv) Ours (STDR+Semi): we combine the STDR strategy with our semi-supervised pipeline.

As shown in Tables~\ref{tab8} and~\ref{tab9}, both STDR-$\alpha$ and STDR-$\beta$ achieve good performance in the two adaptation settings, and in comparison, STDR-$\beta$ outperform, which illustrates that representative samples of the target domain are necessary to significantly improve the performance. Nevertheless, when comparing STDR-$\beta$ with STDR, the experimental outcomes consistently indicate that STDR outperforms overall. This suggests that, for further enhancing model performance, relying solely on domain-specific representative samples is insufficient. It's essential to incorporate domain-invariant samples, enabling more effective fine-tuning of model parameters, preservation of shared segmentation knowledge, and improvement in model generalization, leading to superior results. This integrated approach is crucial for achieving better outcomes. Finally, it's evident that the performance of Ours (STDR+Semi) consistently surpasses that of STDR across nearly all metrics, providing compelling evidence for the efficacy of integrating the semi-supervised pipeline into the active domain adaptation process. 

To assess the robustness of the STDR strategy, we randomly select two adaptation settings and carried out comparative experiments using various proportions of active samples, as shown in Table~\ref{tab10}. To eliminate any potential biases arising from unlabeled samples and the usage of semi-supervised learning methods, we exclusively employed actively labeled target samples during the experiment. As depicted in Table~\ref{tab10}, the model's performance exhibits a consistent improvement with the increase in the proportion of samples. We observe a significant performance boost when introducing new actively labeled (AL) samples, especially with a small sample size. However, this performance improvement gradually levels off as the number of samples increases. Thus, our choice of 20\% of the samples is a trade-off between labeling cost and segmentation performance. Compared to fully labeling all data, our performance is only 4.76\% and 2.34\% lower in DSC for $\mathrm{SPH} \rightarrow \mathrm{SCH}$ and $\mathrm{SPH} \rightarrow \mathrm{APH}$, respectively.

\begin{table}[t]
\centering
\caption{Experimental results  with different numbers of active samples.}
\vspace{-0.6mm}
\label{tab10}
\resizebox{0.5\textwidth}{!}{%
\begin{tabular}{c|c| c c c }
\hline
Adaptation setting & Percent &DSC ($\%$) & HD95 ($mm$) & ASD ($mm$)  \\
\hline
$\mathrm{SPH} \rightarrow \mathrm{SCH}$ & 10\% & $64.61\pm 0.09$ & $11.53\pm 4.21$ & $2.96\pm 1.67$   \\
$\mathrm{SPH} \rightarrow \mathrm{SCH}$ & 20\% & $69.55\pm 0.07$ & $10.64\pm 3.36$ & $2.83\pm 1.83$   \\
$\mathrm{SPH} \rightarrow \mathrm{SCH}$ & 40\% & $70.88\pm 0.07$ & $9.96\pm 4.49$ & $2.47\pm 1.21$ \\
$\mathrm{SPH} \rightarrow \mathrm{SCH}$ & 60\% & $71.73\pm 0.07$ & $9.22\pm 4.28$ & $2.66\pm 1.62$ \\
$\mathrm{SPH} \rightarrow \mathrm{SCH}$ & 80\% & $73.57\pm 0.06$ & $8.82\pm 4.29$ & $2.27\pm 1.19$  \\
$\mathrm{SPH} \rightarrow \mathrm{SCH}$ & 100\% & $74.31\pm 0.06$ & $8.57\pm 4.68$ & $2.09\pm 1.01$  \\
\hline
$\mathrm{SPH} \rightarrow \mathrm{APH}$ & 10\% & $84.05\pm 0.06$ & $5.37\pm 4.24$ & $1.48\pm 1.35$ \\
$\mathrm{SPH} \rightarrow \mathrm{APH}$ & 20\% & $85.05\pm 0.05$ & $4.98\pm 3.07$ & $1.36\pm 1.01$ \\
$\mathrm{SPH} \rightarrow \mathrm{APH}$ & 40\% & $85.62\pm 0.05$ & $4.98\pm 3.13$ & $1.35\pm 1.07$ \\
$\mathrm{SPH} \rightarrow \mathrm{APH}$ & 60\% & $86.02\pm 0.04$ & $4.60\pm 2.47$ & $1.33\pm 0.87$ \\
$\mathrm{SPH} \rightarrow \mathrm{APH}$ & 80\% & $86.81\pm 0.04$ & $4.05\pm 1.99$ & $1.18\pm 0.93$ \\
$\mathrm{SPH} \rightarrow \mathrm{APH}$ & 100\% & $87.39\pm 0.05$ & $3.97\pm 2.19$ & $0.99\pm 0.75$  \\
\hline
\end{tabular}}
\vspace{-5mm}
\end{table}

\section{Disccusion}
Accurate GTV segmentation is pivotal for effective radiotherapy in patients with NPC \cite{xia2000comparison,ng2022current}. Despite the proliferation of GTV segmentation algorithms \cite{li2022npcnet,zhou2006nasopharyngeal,huang2013region,men2017deep,chen2020mmfnet,guo2020automatic}, their practical implementation faces a significant challenge: models trained in the source domain experience substantial performance degradation when applied in a new medical center. Combining Tables~\ref{tab2} and~\ref{tab3}, it's clear that the algorithms consistently exhibit a reduction of 5\% to 10\%, or even more than 10\%, in terms of DSC. There is also a noticeable degradation in performance on HD95 and ASD metrics. For example, in adaptations like $\mathrm{SMU} \rightarrow \mathrm{SCH}$ or $\mathrm{SPH} \rightarrow \mathrm{SCH}$, HD95 shows an error variation exceeding 20$mm$, indicating significant performance discrepancies. Furthermore, the privacy and security of medical data, coupled with hospital regulations, create significant constraints. Even in cases where certain medical centers possess ample well-labeled data, others can't leverage this resource, primarily due to the prominent domain gap evident between different data domains. This disparity is visually represented in Fig.~\ref{fig:tsne}. 

These challenges highlight the need for innovative domain adaptation techniques to bridge this gap, and our proposed solution is Source-Free Active Domain
Adaptation. Our approach holds multiple advantages over existing domain adaptation methods. Firstly, it demands only a limited number of reference vectors, bypassing the need for source data access. This not only ensures data privacy and security but also facilitates easy transmission and utilization. Secondly, our approach offers greater clinical utility. As evident in Tables~\ref{tab4} and~\ref{tab5}, previous methods indeed enhance segmentation model performance on the target domain, but they exclusively rely on unlabeled target data. The absence of ground truth for supervised training can introduce model distortions, resulting in relatively modest improvements, typically in the range of 1-3\% for the DSC. Indeed, within a real clinical context, the active labeling of small data sets is both practically feasible and reasonably efficient in terms of time. The major significance of our approach becomes apparent as its ability to harness this limited data, enabling us to achieve a performance level comparable to that of a fully supervised-trained model specific to the target domain. This underscores the tremendous value our method contributes to the clinical deployment of computer-aid-segmentation networks. For instance, combining Tables~\ref{tab2},~\ref{tab4}, and~\ref{tab5}, we can find that in the $\mathrm{SPH} \rightarrow \mathrm{SCH}$, our method's performance (70.99\%, 8.57$mm$, 2.03$mm$ for DSC, HD95, ASD) is comparable with the results achieved by directly training on SCH (72.19\%, 9.25$mm$, 2.22$mm$ for DSC, HD95, ASD), albeit with a slightly lower DSC and superior HD95 and ASD values. Meanwhile, in the $\mathrm{SMU} \rightarrow \mathrm{SCH}$, our method (72.58\%, 8.06$mm$, 2.00$mm$ for DSC, HD95, ASD) clearly outperforms. 

We recognize certain limitations in our study. Both the development and validation of our deep learning model are conducted exclusively using MRI data due to its superior soft-tissue contrast. While CT is more commonly employed in treatment planning, it's common practice to generate GTVs on MRI and subsequently map them to planning CT through image fusion. However, it would be more clinically pertinent if validation could encompass both MRI and CT datasets.

Another substantial hurdle in this research realm is the scarcity of suitably annotated multi-center GTV datasets. To the best of our knowledge, a major portion of MRI-based GTV segmentation relies on proprietary datasets, and the availability of multi-center datasets remains limited. In response to this challenge, we've taken proactive measures to curate meticulously annotated multi-center GTV datasets, collaborating closely with radiologists. These newly proposed datasets are designed to align with the analytical and practical prerequisites of domain adaptation research. Importantly, we will release an open-source NPC GTV dataset with more than 150 patients from multiple hospitals after the peer review, thereby fostering further progress in the field and encouraging active participation from fellow researchers.

\section{Conclusion}
In this paper, we introduce for the first time a novel Source-Free Active Domain Adaptation for GTV segmentation in cross-center NPC, rigorously validated with data from 1,057 patients at five medical centers. We devise an STDR strategy to discern representative samples of domain-invariant and domain-specific in the target medical center. This selection process hinges on the embedding distance within the latent space, and training with these actively labeled samples significantly improves the model's performance on the target domain. Furthermore, we develop a semi-supervised learning process that combines it with active domain adaptation to fully utilize the remaining unlabeled samples to enhance segmentation. Numerous experimental results show that our network clearly outperforms SOTA methods. Simultaneously, our approach achieves performance near fully supervised training with minimal labeling, showcasing significant clinical medical utility value. In the future, we plan to consider validating our method on other imaging modalities and other tumors.

\bibliographystyle{IEEEtran}
\bibliography{ref}

\begin{thebibliography}{10}
\providecommand{\url}[1]{#1}
\csname url@samestyle\endcsname
\providecommand{\newblock}{\relax}
\providecommand{\bibinfo}[2]{#2}
\providecommand{\BIBentrySTDinterwordspacing}{\spaceskip=0pt\relax}
\providecommand{\BIBentryALTinterwordstretchfactor}{4}
\providecommand{\BIBentryALTinterwordspacing}{\spaceskip=\fontdimen2\font plus
\BIBentryALTinterwordstretchfactor\fontdimen3\font minus \fontdimen4\font\relax}
\providecommand{\BIBforeignlanguage}[2]{{%
\expandafter\ifx\csname l@#1\endcsname\relax
\typeout{** WARNING: IEEEtran.bst: No hyphenation pattern has been}%
\typeout{** loaded for the language `#1'. Using the pattern for}%
\typeout{** the default language instead.}%
\else
\language=\csname l@#1\endcsname
\fi
#2}}
\providecommand{\BIBdecl}{\relax}
\BIBdecl

\bibitem{chua2016nasopharyngeal}
M.~L. Chua, J.~T. Wee \emph{et~al.}, ``Nasopharyngeal carcinoma,'' \emph{The Lancet}, vol. 387, no. 10022, pp. 1012--1024, 2016.

\bibitem{lee2002intensity}
N.~Lee, P.~Xia, J.~M. Quivey, K.~Sultanem, I.~Poon, C.~Akazawa, P.~Akazawa, V.~Weinberg, and K.~K. Fu, ``Intensity-modulated radiotherapy in the treatment of nasopharyngeal carcinoma: an update of the ucsf experience,'' \emph{IJROBP}, vol.~53, no.~1, pp. 12--22, 2002.

\bibitem{luo2023deep}
X.~Luo, W.~Liao, Y.~He \emph{et~al.}, ``Deep learning-based accurate delineation of primary gross tumor volume of nasopharyngeal carcinoma on heterogeneous magnetic resonance imaging: A large-scale and multi-center study,'' \emph{Radiotherapy and Oncology}, vol. 180, p. 109480, 2023.

\bibitem{chen2019nasopharyngeal}
Y.-P. Chen, A.~T. Chan, Q.-T. Le \emph{et~al.}, ``Nasopharyngeal carcinoma,'' \emph{The Lancet}, vol. 394, no. 10192, pp. 64--80, 2019.

\bibitem{kam2004treatment}
M.~K. Kam, P.~M. Teo, R.~M. Chau \emph{et~al.}, ``Treatment of nasopharyngeal carcinoma with intensity-modulated radiotherapy: the hong kong experience,'' \emph{IJROBP}, vol.~60, no.~5, pp. 1440--1450, 2004.

\bibitem{razek2012mri}
A.~A. K.~A. Razek and A.~King, ``Mri and ct of nasopharyngeal carcinoma,'' \emph{AJOR}, vol. 198, no.~1, pp. 11--18, 2012.

\bibitem{lin2019deep}
L.~Lin, Q.~Dou, Y.-M. Jin \emph{et~al.}, ``Deep learning for automated contouring of primary tumor volumes by mri for nasopharyngeal carcinoma,'' \emph{Radiology}, vol. 291, no.~3, pp. 677--686, 2019.

\bibitem{chen2022failure}
S.~Chen, D.~Yang, X.~Liao \emph{et~al.}, ``Failure patterns of recurrence and metastasis after intensity-modulated radiotherapy in patients with nasopharyngeal carcinoma: results of a multicentric clinical study,'' \emph{Frontiers in Oncology}, vol.~11, p. 5730, 2022.

\bibitem{wang2020boundary}
S.~Wang, M.~Liu, J.~Lian, and D.~Shen, ``Boundary coding representation for organ segmentation in prostate cancer radiotherapy,'' \emph{TMI}, vol.~40, no.~1, pp. 310--320, 2020.

\bibitem{jin2021deeptarget}
D.~Jin, D.~Guo \emph{et~al.}, ``Deeptarget: Gross tumor and clinical target volume segmentation in esophageal cancer radiotherapy,'' \emph{MedIA}, vol.~68, p. 101909, 2021.

\bibitem{tian2023delineation}
M.~Tian, H.~Wang, X.~Liu \emph{et~al.}, ``Delineation of clinical target volume and organs at risk in cervical cancer radiotherapy by deep learning networks,'' \emph{Medical Physics}, 2023.

\bibitem{li2022npcnet}
Y.~Li, T.~Dan \emph{et~al.}, ``Npcnet: jointly segment primary nasopharyngeal carcinoma tumors and metastatic lymph nodes in mr images,'' \emph{TMI}, vol.~41, no.~7, pp. 1639--1650, 2022.

\bibitem{weiss2016survey}
K.~Weiss, T.~M. Khoshgoftaar, and D.~Wang, ``A survey of transfer learning,'' \emph{Journal of Big data}, vol.~3, no.~1, pp. 1--40, 2016.

\bibitem{wang2020tent}
D.~Wang, E.~Shelhamer, S.~Liu, B.~Olshausen, and T.~Darrell, ``Tent: Fully test-time adaptation by entropy minimization,'' \emph{ICLR}, 2021.

\bibitem{zhou2006nasopharyngeal}
J.~Zhou, K.~L. Chan, P.~Xu, and V.~F. Chong, ``Nasopharyngeal carcinoma lesion segmentation from mr images by support vector machine,'' in \emph{ISBI}, 2006, pp. 1364--1367.

\bibitem{huang2013region}
W.~Huang, K.~L. Chan, and J.~Zhou, ``Region-based nasopharyngeal carcinoma lesion segmentation from mri using clustering-and classification-based methods with learning,'' \emph{JDI}, vol.~26, pp. 472--482, 2013.

\bibitem{men2017deep}
K.~Men, X.~Chen \emph{et~al.}, ``Deep deconvolutional neural network for target segmentation of nasopharyngeal cancer in planning computed tomography images,'' \emph{Frontiers in oncology}, vol.~7, p. 315, 2017.

\bibitem{chen2020mmfnet}
H.~Chen, Y.~Qi, Y.~Yin \emph{et~al.}, ``Mmfnet: A multi-modality mri fusion network for segmentation of nasopharyngeal carcinoma,'' \emph{Neurocomputing}, vol. 394, pp. 27--40, 2020.

\bibitem{liao2022automatic}
W.~Liao, J.~He \emph{et~al.}, ``Automatic delineation of gross tumor volume based on magnetic resonance imaging by performing a novel semisupervised learning framework in nasopharyngeal carcinoma,'' \emph{IJROBP}, vol. 113, no.~4, pp. 893--902, 2022.

\bibitem{wilson2020survey}
G.~Wilson and D.~J. Cook, ``A survey of unsupervised deep domain adaptation,'' \emph{TIST}, vol.~11, no.~5, pp. 1--46, 2020.

\bibitem{tsai2018learning}
Y.-H. Tsai, W.-C. Hung \emph{et~al.}, ``Learning to adapt structured output space for semantic segmentation,'' in \emph{CVPR}, 2018, pp. 7472--7481.

\bibitem{vu2019advent}
T.-H. Vu, H.~Jain, M.~Bucher, M.~Cord, and P.~P{\'e}rez, ``Advent: Adversarial entropy minimization for domain adaptation in semantic segmentation,'' in \emph{CVPR}, 2019, pp. 2517--2526.

\bibitem{guan2021domain}
H.~Guan and M.~Liu, ``Domain adaptation for media: a survey,'' \emph{TBME}, vol.~69, no.~3, pp. 1173--1185, 2021.

\bibitem{fleuret2021uncertainty}
F.~Fleuret \emph{et~al.}, ``Uncertainty reduction for model adaptation in semantic segmentation,'' in \emph{CVPR}, 2021, pp. 9613--9623.

\bibitem{chen2021source}
C.~Chen, Q.~Liu, Y.~Jin, Q.~Dou, and P.-A. Heng, ``Source-free domain adaptive fundus image segmentation with denoised pseudo-labeling,'' in \emph{MICCAI}.\hskip 1em plus 0.5em minus 0.4em\relax Springer, 2021, pp. 225--235.

\bibitem{cohn1996active}
D.~A. Cohn, Z.~Ghahramani, and M.~I. Jordan, ``Active learning with statistical models,'' \emph{JAIR}, vol.~4, pp. 129--145, 1996.

\bibitem{lewis1994heterogeneous}
D.~D. Lewis and J.~Catlett, ``Heterogeneous uncertainty sampling for supervised learning,'' in \emph{ML}.\hskip 1em plus 0.5em minus 0.4em\relax Elsevier, 1994, pp. 148--156.

\bibitem{scheffer2001active}
T.~Scheffer, C.~Decomain, and S.~Wrobel, ``Active hidden markov models for information extraction,'' in \emph{ISIDA}.\hskip 1em plus 0.5em minus 0.4em\relax Springer, 2001, pp. 309--318.

\bibitem{jain2016active}
S.~D. Jain and K.~Grauman, ``Active image segmentation propagation,'' in \emph{CVPR}, 2016, pp. 2864--2873.

\bibitem{hoi2009semisupervised}
S.~C. Hoi, R.~Jin, J.~Zhu, and M.~R. Lyu, ``Semisupervised svm batch mode active learning with applications to image retrieval,'' \emph{TOIS}, vol.~27, no.~3, pp. 1--29, 2009.

\bibitem{xu2003representative}
Z.~Xu, K.~Yu, V.~Tresp, X.~Xu, and J.~Wang, ``Representative sampling for text classification using support vector machines,'' in \emph{ECIR}.\hskip 1em plus 0.5em minus 0.4em\relax Springer, 2003, pp. 393--407.

\bibitem{huang2010active}
S.-J. Huang, R.~Jin, and Z.-H. Zhou, ``Active learning by querying informative and representative examples,'' \emph{NeurIPS}, vol.~23, 2010.

\bibitem{freytag2014selecting}
A.~Freytag, E.~Rodner, and J.~Denzler, ``Selecting influential examples: Active learning with expected model output changes,'' in \emph{ECCV}.\hskip 1em plus 0.5em minus 0.4em\relax Springer, 2014, pp. 562--577.

\bibitem{kading2015active}
C.~Kading, A.~Freytag, E.~Rodner, P.~Bodesheim, and J.~Denzler, ``Active learning and discovery of object categories in the presence of unnameable instances,'' in \emph{CVPR}, 2015, pp. 4343--4352.

\bibitem{su2020active}
J.-C. Su, Y.-H. Tsai, K.~Sohn, B.~Liu, S.~Maji, and M.~Chandraker, ``Active adversarial domain adaptation,'' in \emph{WACV}, 2020, pp. 739--748.

\bibitem{hinton2002stochastic}
G.~E. Hinton and S.~Roweis, ``Stochastic neighbor embedding,'' \emph{NeurIPS}, vol.~15, 2002.

\bibitem{macqueen1967some}
J.~MacQueen \emph{et~al.}, ``Some methods for classification and analysis of multivariate observations,'' in \emph{symposium on mathematical statistics and probability}, vol.~1, no.~14.\hskip 1em plus 0.5em minus 0.4em\relax Oakland, CA, USA, 1967, pp. 281--297.

\bibitem{ronneberger2015u}
O.~Ronneberger, P.~Fischer, and T.~Brox, ``U-net: Convolutional networks for biomedical image segmentation,'' in \emph{MICCAI}, 2015, pp. 234--241.

\bibitem{wang2023pymic}
G.~Wang, X.~Luo, R.~Gu, S.~Yang, Y.~Qu, S.~Zhai, Q.~Zhao, K.~Li, and S.~Zhang, ``Pymic: A deep learning toolkit for annotation-efficient medical image segmentation,'' \emph{CMPB}, vol. 231, p. 107398, 2023.

\bibitem{xia2000comparison}
P.~Xia, K.~K. Fu \emph{et~al.}, ``Comparison of treatment plans involving intensity-modulated radiotherapy for nasopharyngeal carcinoma,'' \emph{IJROBP}, vol.~48, no.~2, pp. 329--337, 2000.

\bibitem{ng2022current}
W.~T. Ng, J.~C. Chow \emph{et~al.}, ``Current radiotherapy considerations for nasopharyngeal carcinoma,'' \emph{Cancers}, vol.~14, no.~23, p. 5773, 2022.

\bibitem{guo2020automatic}
Y.~Guo, Q.~Yang, W.~Hu, Z.~Zhang, J.~Wang, and C.~Hu, ``Automatic segmentation of nasopharyngeal carcinoma on mr images: A single-institution experience,'' \emph{IJROBP}, vol. 108, no.~3, p. e776, 2020.

\end{thebibliography}

\end{document}